\documentclass[10pt,journal,compsoc]{IEEEtran}
\ifCLASSOPTIONcompsoc
  \usepackage[nocompress]{cite}
\else
  \usepackage{cite}
\fi
\usepackage{amssymb}
\usepackage{pifont}
\usepackage{threeparttable}

\ifCLASSINFOpdf
   \usepackage[pdftex]{graphicx}
\else
\fi
\usepackage{amsmath}
\usepackage{multirow}
\usepackage{booktabs}

\usepackage{array}
\begin{document}
%
\title{Geo-NI: Geometry-aware Neural Interpolation for Light Field Rendering}
%
%
%
%
\author{Gaochang~Wu,
		Yuemei~Zhou,
		Lu~Fang,~\IEEEmembership{Senior Member,~IEEE,}
		Yebin~Liu,~\IEEEmembership{Member,~IEEE,}
        and~Tianyou~Chai,~\IEEEmembership{Life Fellow,~IEEE}
\IEEEcompsocitemizethanks{\IEEEcompsocthanksitem Gaochang Wu and Tianyou Chai are with the State Key Laboratory of Synthetical Automation for Process Industries, Northeastern University, Shenyang 110819, P. R. China. E-mail: \{wugc, tychai\}@mail.neu.edu.cn.
\IEEEcompsocthanksitem Yuemei Zhou and Yebin Liu are with Department of Automation, Tsinghua University, Beijing 100084, P. R. China. Email: zym18@mails.tsinghua.edu.cn, liuyebin@mail.tsinghua.edu.cn.
\IEEEcompsocthanksitem Fang Lu is with the Department of Electronic Engineering, Tsinghua University, Beijing 100084, China, and also with the Beijing National Research Center for Information Science and Technology, Beijing 100084, P. R. China. Email: fanglu@tsinghua.edu.cn.}

\thanks{Corresponding Author: Tianyou Chai.}
}

%
%

\markboth{}%
{Shell \MakeLowercase{\textit{et al.}}: Bare Demo of IEEEtran.cls for Computer Society Journals}
%



\IEEEtitleabstractindextext{%
\begin{abstract}
We present a novel Geometry-aware Neural Interpolation (Geo-NI) framework for light field rendering. Previous learning-based approaches either perform direct interpolation via neural networks, which we dubbed Neural Interpolation (NI), or explore scene geometry for novel view synthesis, also known as Depth Image-Based Rendering (DIBR). Both kinds of approaches have their own strengths and weaknesses in addressing non-Lambert effect and large disparity problems. In this paper, we incorporate the ideas behind these two kinds of approaches by launching the NI within a specific DIBR pipeline. Specifically, a DIBR network in the proposed Geo-NI serves to construct a novel reconstruction cost volume for neural interpolated light fields sheared by different depth hypotheses. The reconstruction cost can be interpreted as an indicator reflecting the reconstruction quality under a certain depth hypothesis, and is further applied to guide the rendering of the final high angular resolution light field. To implement the Geo-NI framework more practically, we further propose an efficient modeling strategy to encode high-dimensional cost volumes using a lower-dimension network. By combining the superiorities of NI and DIBR, the proposed Geo-NI is able to render views with large disparities with the help of scene geometry while also reconstructing the non-Lambertian effect when depth is prone to be ambiguous. Extensive experiments on various datasets demonstrate the superior performance of the proposed geometry-aware light field rendering framework.
\end{abstract}

}

\maketitle

\IEEEdisplaynontitleabstractindextext

%
\IEEEpeerreviewmaketitle

\IEEEraisesectionheading{\section{Introduction}}\label{sec:introduction}

%
%
%
%
\IEEEPARstart{L}{ight} field (LF) describes rays travelling from all directions in a free space~\cite{LFrendering}, demultiplexing the angular information lost in conventional 2D imaging~\cite{Wu2022AnII}. Benefits from the LF rendering technologies~\cite{LFrendering,gortler1996lumigraph}, LF enables to reproduce photorealistic views in real-time, enabling travelling freely in metaverse. Standard LF rendering technologies require a Nyquist rate view sampling, i.e., densely-sampled LF with disparities between adjacent views to be less than one pixel~\cite{Lin2004Geo}. However, existing densely-sampled LF devices or systems~\cite{wu2017light} either suffers from a long period of acquisition time or falls into the well-known resolution trade-off problem, i.e., sacrificing the spatial resolution for a dense sampling in the angular domain.

With the success of deep learning in artificial intelligence~\cite{David2017Mastering}, recent researches~\cite{DoubleCNN,WuEPICNN2018,LFrig,mildenhall2019local} are stepping towards deep learning-based interpolation, which we refer to as Neural Interpolation (NI), or Depth Image-Based Rendering (DIBR) using a sparsely-sampled LF in the angular domain. On the one hand, typical learning-based NI methods~\cite{LFCNN,WuEPICNN2018,YeungECCV2018} directly map the low angular resolution LF to densely-sampled LF through diverse network architectures. These methods are highly effective in modelling non-Lambertian effects. Nevertheless, the perception range (receptive field) of the network~\cite{Long2014Do,zhou2015object} limits the performance on LF with large disparities, leading to aliasing effects in the reconstructed LF. On the other hand, state-of-the-art learning-based DIBR methods~\cite{DoubleCNN,mildenhall2019local,jin2020light} resort to depth estimation followed by view synthesis, which is considered to be a more efficient way to deal with the large disparity issue than only relying on the receptive field. But these methods require depth consistency along the angular dimension, and thus, often fail to handle the depth ambiguity caused by the non-Lambertian effect.

\begin{figure}
\begin{center}
\includegraphics[width=1\linewidth]{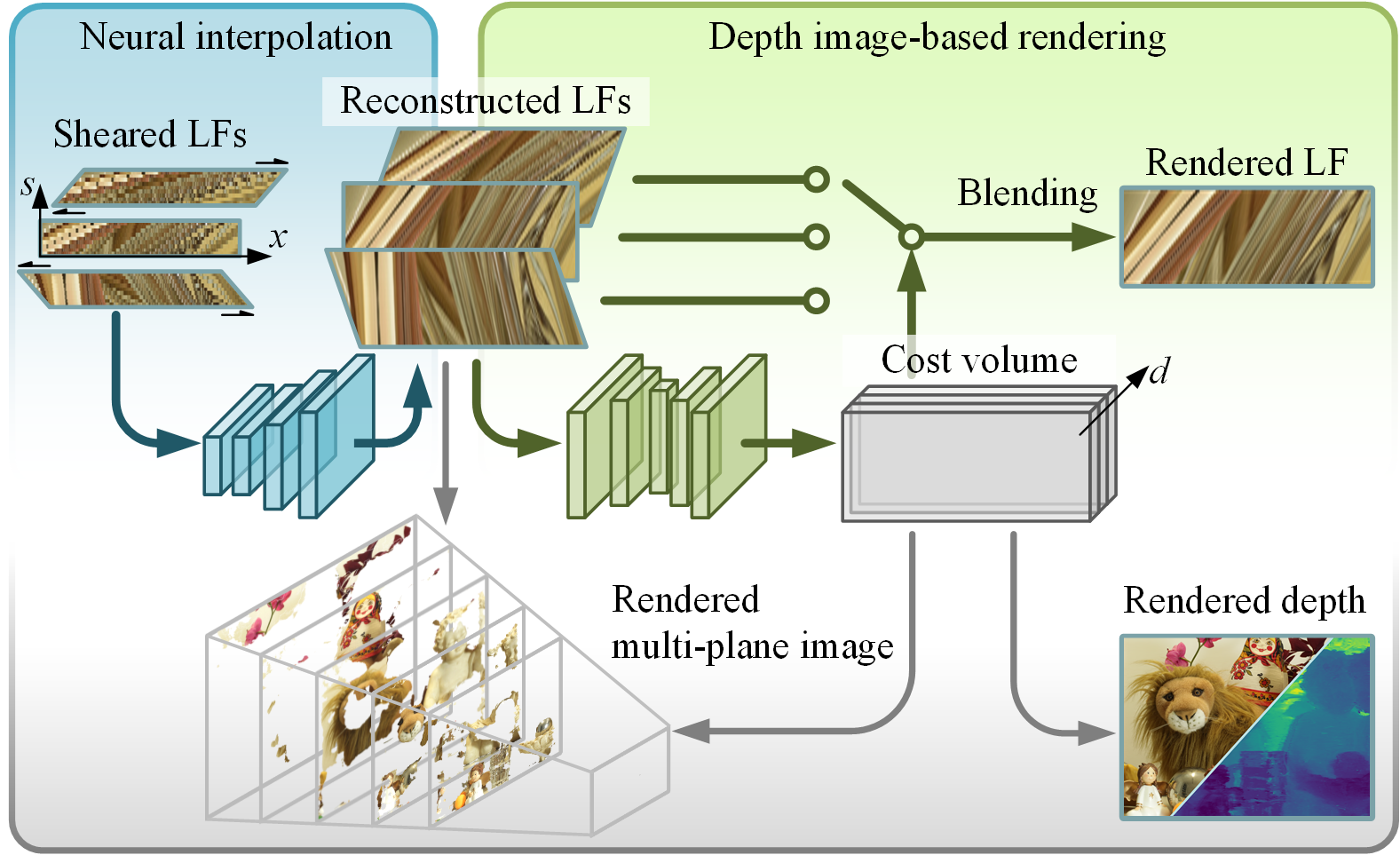}
\end{center}
\vspace{-4mm}
   \caption{We present Geo-NI framework for geometry-aware light field rendering by launching the Neural Interpolation (NI) within a novel Depth Image-Based Rendering (DIBR) pipeline. The proposed framework is able to render LFs with large disparity while also reconstruct the non-Lambertian effects. Due to the awareness of the scene geometry, the proposed framework is able to render multi-plane image (a layered scene representation) and scene depth. LF courtesy of Adhikarla~\textit{et al.}~\cite{kiran2017towards}.}
\label{fig:Teaser}
\vspace{-2mm}
\end{figure}

In this paper, we propose a learning-based framework for geometry-aware LF rendering to address the non-Lambertian and large disparity issues by launching the NI within a well-designed DIBR pipeline. We term the framework as Geo-NI, as shown in Fig.~\ref{fig:Teaser}. Specifically, we bridge the gap between the standard NI and DIBR by shearing the input LF with a set of depth hypotheses. The NI part is achieved via a neural network that directly interpolates the sheared LFs. Then the DIBR part is implemented via another neural network to assess the reconstruction quality of the neural interpolated LFs under the depth hypotheses. On this basis, the DIBR part constructs a cost volume, where each value in the volume can be interpreted as a weight for blending the final rendered (output) LF, as shown in Fig.~\ref{fig:Teaser}. We, therefore, name the volume reconstruction cost volume. This feature ensures the interpretability of the overall framework by explicitly showing the depth hypothesis chosen by the network, i.e., the geometry awareness. Moreover, the proposed reconstruction cost volume can be converted to high-quality depth and multi-plane image (a layered scene representation)~\cite{zhou2018stereo}, as shown in Fig.~\ref{fig:Teaser} (bottom). Please refer to Sec.~\ref{Sec:analysis} for details.

Directly modeling the reconstruction cost volume suffers the curse of dimensionality due to introducing an additional (shear) dimension for LFs, which inevitably resulting in redundant parameters in the network. To tackle this problem, we consider each slice in the cost volume to be independent of each other along the shear dimension. This allows us to construct the high-dimensional cost volume with a network of lower dimensional and lower computational complexity. We further propose a novel hierarchical packing-unpacking structure that is able to efficiently encode and decode the spatial-angular features of each slice for the LF reconstruction in the NI part and cost construction in the DIBR part. The building block of the packing/unpacking structure is spatial-to-channel/channel-to-spatial pixel shuffling followed by convolutional layers. With the help of the spatial-channel pixel shuffling, the networks in Geo-NI are able to efficiently gain a large perceptive field on the spatial-angular dimensions by intactly compressing the spatial resolution and restore it without high-frequency loss. In summary, we make the following  contributions: 
\begin{itemize}
    \item An interpretable Geometry-aware Neural Interpolation (Geo-NI) framework that joints neural interpolation and depth-based view synthesis for solving non-Lambertian and large disparity challenges in an end-to-end manner;
    \item A well-designed reconstruction cost volume derived from the DIBR pipeline that guides the blending of the LFs sheared by different depth hypotheses. Even in the absence of depth supervision, the reconstruction cost volume can be applied for rendering high-quality depth and multi-plane images;
    \item An efficient modeling strategy for high-dimensional reconstruction cost volume using lower-dimensional network. The network infers the cost volume slice by slice along the shear dimension and encodes the spatial-angular features of LFs with hierarchical packing-unpacking structure.
\end{itemize}

We demonstrate the superiority of the proposed Geo-NI framework by performing extensive evaluations on various LF datasets. By incorporating the NI and DIBR parts, the proposed Geo-NI framework presents high-quality results on challenging cases with large disparities while also reconstruct the non-Lambertian effect when depth is prone to be ambiguous.

\section{Related Work}
\subsection{Plenoptic Sampling and Reconstruction} 
These approaches treat LF reconstruction as the approximation of plenoptic function using a set of samples. The analysis tools in the Fourier domain show that the sampling produces spectrum replicas along the sample dimensions. Under sparse sampling, the replicas will overlap with their original spectrum, resulting in aliasing effects in the LF signals. Classical approaches by Chai \textit{et al.}~\cite{chai2000plenoptic} and Zhang \textit{et al.}~\cite{zhang2003spectral} formulate the reconstruction as filtering of the aliasing high-frequencies while keeping the original spectrum as complete as possible. In~\cite{Shearlet}, Vagharshakyan \textit{et al.} explored a shearlet transform of composite directions and scales in the Fourier domain to remove the aliasing high-frequencies. To handle the occlusion problem, Zhu~\textit{et al.}~\cite{Zhu20Signal} proposed an occlusion field theory to quantify the occlusion degree and designed a reconstruction filter to compensate for the missing information caused by the occlusion.

Researchers also focus on training a deep neural network to directly infer the densely-sampled LF from the low angular resolution input~\cite{LFCNN,WuEPICNN2018,YeungECCV2018,zhu2019revisiting}. We term such approaches as Neural Interpolation (NI) since they regard the reconstruction as a learning-based interpolation problem of the missing pixels (samples) without exploiting scene geometry. An initial work by Yoon~\emph{et al}.~\cite{LFCNN} first interpolates the low angular resolution LF with the bicubic operation, then employs a convolutional network to refine the reconstructed LF. For explicitly handling the aliasing effects, Wu \emph{et al}.~\cite{WuEPICNN2018} proposed a ``blur-restoration-deblur'' framework that first suppresses the high-frequency components in the spatial dimension and then restores them via a non-blind deconvolution. Li \emph{et al}.~\cite{Li2023DenseLF} proposed a novel epipolar focus spectrum representation, and applied a network to eliminate the aliasing frequencies in the Fourier domain. Wang \textit{et al.}~\cite{wang2020high} applied 3D convolution layers to reconstruct the two angular dimensions of the input LF sequentially. However, the small perception range, i.e., receptive field~\cite{Long2014Do,zhou2015object}, impedes the networks to capture long-term correspondences in the input LF, resulting in limited performances. 

Modeling light fields with a deeper network or an efficient architecture to enlarge the receptive field can significantly improve the reconstruction performance. Yeung \textit{et al.}~\cite{YeungECCV2018} directly fed the entire 4D LF into a pseudo 4D convolutional network and proposed a novel spatial-angular alternating convolution to iteratively refine the angular dimensions of the LF. Jin \textit{et al.}~\cite{jin2020light} further extended the spatial-angular alternating convolution to the problem of compressive LF reconstruction. Zhu \emph{et al}.~\cite{zhu2019revisiting} introduced a U-net architecture to enlarge the receptive field using strided convolutional layers and convLSTM layers~\cite{shi2015convolutional}. Liu \textit{et al.}~\cite{liu2020multi}, Zhang \textit{et al.}~\cite{Zhang21Micro} and Meng \textit{et al.}~\cite{meng2019high} applied residual blocks~\cite{he2016deep} and dense blocks~\cite{Huang_2017_CVPR} to prevent gradient vanishment when increasing the depth of the networks. Wang~\textit{et al.}~\cite{ying2021disentangling} proposed an LF disentangling mechanism to disentangle the coupled spatial-angular information by using dilated and strided convolutions directly on the macro-pixel image. However, since the actual size of the receptive field can be smaller than its theoretical size~\cite{zhou2015object}, simply pursuing a deeper network without modeling the scene geometry still limits the performance of NI-based approaches.

A recent research direction is encoding the 5D plenoptic function within a coordinate-based neural network, which is termed Neural Radiance Field (NeRF) representation. The pioneering work by Mildenhall~\textit{et al.}~\cite{mildenhall2020nerf} optimizes a network with 5D samples (the 3D location $(x,y,z)$ as well as 2D view direction $(\theta,\phi)$) individually for each scene. To achieve the generalization to arbitrary scenes, Wang~\textit{et al.}~\cite{Wang_2021_CVPR} presented IBRNet to predict colors weighted by features from neighboring views. Chen~\textit{et al.}~\cite{Chen2021MVSNeRFFG} proposed a neural scene encoding volume using plane sweep input and decode the volume density and radiance at arbitrary view directions. Suhail~\textit{et al.}~\cite{Suhail_2022_CVPR} further developed a light field-based NeRF that employs transformer-based ray fusion within the constraint of epipolar geometry.

\begin{figure*}
	\begin{center}
		\includegraphics[width=1\linewidth]{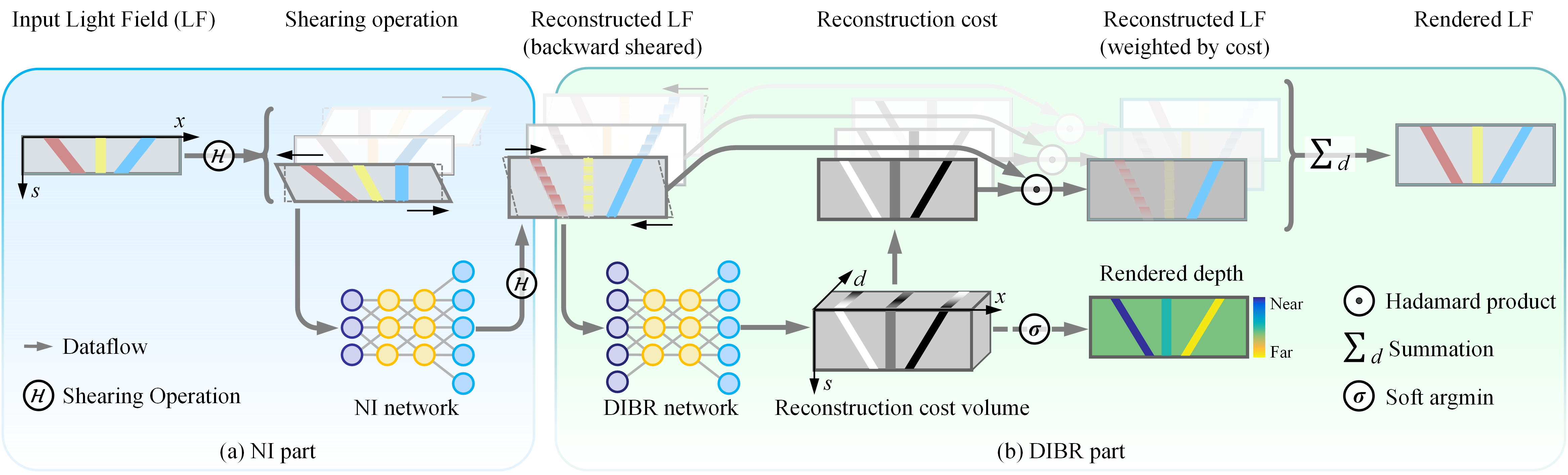}
	\end{center}
	\vspace{-4mm}
	\caption{The proposed Geo-NI framework is composed of two parts: (a) a Neural Interpolation (NI) part that directly reconstructs the sheared LF (Sec. \ref{Sec:network}); and (b) a Depth Image-Based Rendering (DIBR) part that assigns a reconstruction cost map to each reconstructed LF (Sec. \ref{Sec:network}). The DIBR part then blend the reconstructed LFs according to the reconstruction cost. The input is a 3D slice ($L(u,v,s)$ or $L(v,u,t)$) of the LF. We visualize the slices as 2D EPIs for better understanding.}
	\label{fig:CNN}
\end{figure*}

\subsection{Depth Image-based Rendering} 
These approaches first estimate the scene geometry (or depth), then warp the input images to the target viewpoint according to the estimated geometry and blend them. Standard LF depth estimation approaches follow the pipeline of stereo matching~\cite{scharstein2002taxonomy}, which consists of feature description (extraction), cost computation, cost aggregation (or cost volume filtering), depth regression, and post refinement. Because of the difference in the data attribute, LF provides various depth cues for the feature description and cost computation, e.g., structure tensor-based local direction estimation~\cite{Wanner}, orthographic Hough transform for curve estimation~\cite{Vianello2018Robust}, depth from correspondence~\cite{huang2017robust}, depth from defocus~\cite{Tao,Occ} and depth from parallelogram cues~\cite{zhang2016robust}. Some recent learning-based approaches also explored the cost aggregation and depth regression in the aforementioned pipeline with 2D or 3D convolution networks~\cite{kendall2017end,tsaiattention}.

For rendering or synthesizing a novel view, typical DIBR approaches first warp the input views to the novel viewpoint with sub-pixel accuracy and then blend them using different strategies, such as total variation optimization~\cite{Wanner}, soft blending~\cite{soft3D}, and learning-based synthesis~\cite{ko2021light,zhou2021cross,Guo_2021_ICCV}. The most representative DIBR approaches~\cite{DeepStereo,DoubleCNN} using deep learning techniques employ a sequential network setting and train the network models by minimizing errors between the synthesized views and the desired outputs (labels). An initial work by Kalantari \textit{et al.}~\cite{DoubleCNN} proposed an end-to-end DIBR framework using two sequential networks to infer depth and color, respectively. Following this setting, Shi \textit{et al.}~\cite{shi2020learning} proposed to render novel views by blending the warped images at both pixel-level and feature-level. Meng~\textit{et al.}~\cite{meng2021light} introduced a network for estimating warping confidences that address the errors around occlusion regions. Jin \textit{et al.} proposed a spatial-angular alternating refinement network for images warped by using a regular sampling pattern (four corner views)~\cite{jin2020learning} and a flexible sampling pattern~\cite{jin2020deep}.

Different from the sequential network settings, recent researches~\cite{zhou2018stereo,John2019DeepView,Han2022SingleViewVS} focus on decomposing input RGB images into depth-dependent layers, which is dubbed Multi-Plane Image (MPI). For example, Zhou \textit{et al.}~\cite{zhou2018stereo} proposed a learning-based framework that infers a single MPI and a background image from two stereo images and synthesizes novel views via alpha blending. Based on this pioneering work, Mildenhall \textit{et al.}~\cite{mildenhall2019local} proposed to infer an MPI for each view in the input LF and synthesize novel views by blending the local MPIs. Instead of storing the static RGB values in the MPI representation, Wizadwongsa~\textit{et al.}~\cite{Wizadwongsa_2021_CVPR} and Phongthawee~\textit{et al.}~\cite{Nex360} model each pixel as a function of the viewing angle approximated by the combination of spherical basis functions.

The aforementioned DIBR approaches directly blend the views warped according to the depth map or depth-related layers, which is highly effective for scenes with large disparity. However, since the depth information is deduced based on the Lambertian assumption, it will appear ambiguity around non-Lambertian regions, resulting in ghosting effects in the synthesized views. In this paper, instead of directly performing pixel or layer-wise warping using depth information, we propose to render the entire high-angular resolution LF by blending the neural interpolated LFs. Due to the NI does not rely on depth information, the proposed Geo-NI framework shows higher reconstruction quality around non-Lambertian regions.

\section{Methodology}\label{Sec:method}
An LF can be parameterized as a 4D function using two-plane representation~\cite{LFrendering}, i.e., $\mathbb{L}(x,y,s,t)\in\mathbb{R}^{W\times H\times A_s\times A_t}$ with $(x,y)$ denoting the spatial plane  and $(s,t)$ the angular plane. A sub-aperture image is obtained by fixing two angular dimensions of the 4D LF $I(x,y)=\mathbb{L}_{s^*,t^*}(x,y)$. An Epipolar Plane Image (EPI) is extracted by fixing one spatial dimension and one angular dimension, $E(x,s)=\mathbb{L}_{y^*,t^*}(x,s)$ or $E(y,t)=\mathbb{L}_{x^*,s^*}(y,t)$. In this paper, we use a 3D LF slice with two spatial dimensions and one angular dimension, i.e., $L(x,y,s)=\mathbb{L}_{t^*}(x,y,s)$ or $L(y,x,t)=\mathbb{L}_{s^*}(x,y,t)$. By splitting LFs into 3D slices, the proposed method can be applied to both 3D LFs from a single-degree-of-freedom gantry system~\cite{kiran2017towards,ICME2018} and 4D LFs from plenoptic camera~\cite{Lytro} or camera array system. For the reconstruction of a 4D LF $\mathbb{L}(x,y,s,t)$, we employ a hierarchical reconstruction strategy introduced in~\cite{WuEPICNN2018}. In the first step, we first reconstruct 3D LFs using slices $\mathbb{L}_{t^*}(x,y,s)$ and $\mathbb{L}_{s^*}(x,y,t)$ as input. We then apply the previously reconstructed 3D slices to synthesize the final 4D LF. 

In this section, we first introduce the Geometry-aware Neural Interpolation (Geo-NI) framework in Sec. \ref{Sec:framework}, then discuss the relation between the Geo-NI and standard DIBR pipeline in Sec. \ref{Sec:relation2DIBR}, and finally present the proposed hierarchical packing-unpacking structure in detail (Sec. \ref{Sec:network}). 

\subsection{Geometry-aware Neural Interpolation Framework}\label{Sec:framework}
In this paper, we hope to design a geometry-aware LF rendering framework that promises to solve the non-Lambertian and large disparity issues in an end-to-end manner. Our proposal is to solve these issues by implanting the characteristic of NI into a DIBR pipeline. This also brings interpretability to the overall framework. The overall framework is depicted in Fig. \ref{fig:CNN}. In the following, we will explain the details of the proposed framework according to the order of the data flow.

\textbf{Shearing operation.} It is an essential module that bridges the NI part and the DIBR part in the proposed framework. The shearing operation directly alters disparity by shifting all the sub-aperture images synchronously~\cite{ren05,Tao}. For an input 3D LF slice $L(x,y,s)$ (or $L(y,x,t)$), we first shear it with a set of depth hypotheses $d\in \mathbb{D}$. The shearing operation can be formulated as follows
\begin{equation}\label{eq:shearing}
\mathcal{H}_d(L(x,y,s))=L(x+(s-\frac{S}{2})\cdot d,y,s),
\end{equation}
where $\mathcal{H}_d(\cdot)$ denotes the shearing operation with depth hypothesis $d$ and $S$ is the angular resolution of the input LF slice. The vanilla implementation in~\cite{ren05,Tao} describes the shearing operation as $L(x+s\cdot d,y,s)$. We add the term $-\frac{S}{2}\cdot d$ to avoid losing too many boundary pixels in the spatial dimension $x$. Besides, we use zero padding for the blank regions caused by the shearing operation in our implementation.

The shearing operation generates a set of sheared LF slices $\{\mathcal{H}_{d}(L(x,y,s))\}_{d\in\mathbb{D}}$, which is termed sheared volume, as indicated in Fig.~\ref{fig:CNN}(a).

\begin{figure}
	\begin{center}
		\includegraphics[width=1\linewidth]{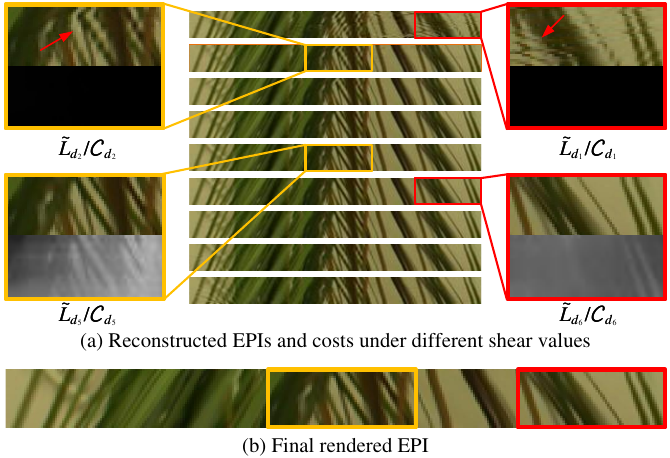}
	\end{center}
	\vspace{-4mm}
	\caption{An illustration of rendering the final high angular resolution LF (EPI) using the reconstructed LFs $\tilde{L}_d$ and their corresponding reconstruction costs $\mathcal{C}_d$. Reconstructed EPIs with improper shear values will appear severe aliasing effects, as shown by the top left and top right zoom-in images. The DIBR network serves to detect the aliasing degree of the reconstructed LFs $\tilde{L}_d$ and predict pixel-wise weights, i.e., reconstruction costs $\mathcal{C}_d$).}
	\label{fig:fusionshear}
\end{figure}

\textbf{Neural interpolation.} We then feed the slices $\mathcal{H}_{d}(L(x,y,s))$ into a neural network, which we dubbed NI network, to directly reconstruct high angular resolution LFs. This step can be denoted as 
$$\bar{L}_{d}(x,y,s)=f_N(\mathcal{H}_{d}(L(x,y,s));\theta),$$
where $f_N(\cdot;\theta)$ is the NI network with trainable parameters $\theta$. Detailed configuration of the NI network will be introduced in Sec. \ref{Sec:network}. Note that the reconstructed LFs after this step are still under the deformation effect of the shearing operation. Therefore, we apply another shearing operation to eliminate this effect after the plenoptic reconstruction, which is termed backward shearing for short. This step can be denoted as
$$\tilde{L}_{d}(x,y,s)=\mathcal{H}_{-\frac{d}{\alpha}}(\bar{L}_d(x,y,s)),$$
where $\alpha$ is the upsampling factor in the angular dimension and $\mathcal{H}$ is the shearing operation described in Eqn.~\ref{eq:shearing}. Due to the reconstruction, the shear amount in the backward shearing operation is $-\frac{d}{\alpha}$ rather than $-d$. Fig. \ref{fig:fusionshear}(a) illustrates reconstructed LFs (EPIs) under a set of shear values. An appropriate shear value will help the NI network generate an aliasing-free high angular resolution LF, while an improper shear value will exacerbate the aliasing effects.

\begin{figure*}
\begin{center}
\includegraphics[width=1.\linewidth]{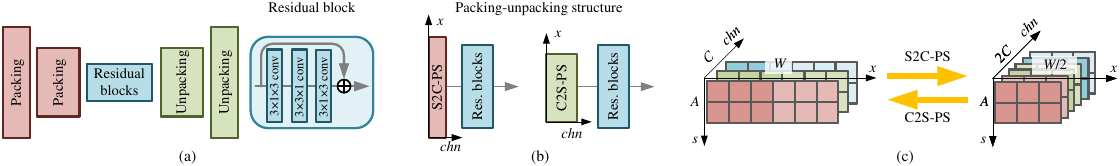}
\end{center}
\vspace{-4mm}
\caption{(a) An illustration of packing-unpacking network; (b) The packing block for the encoder (left) and the unpacking for the block decoder (right); (c) An illustration of Spatial-to-Channel Pixel Shuffle (S2C-PS) and Channel-to-Spatial Pixel Shuffle (C2S-PS).}
\label{fig:pixelshuffle}
\end{figure*}

\textbf{Reconstruction cost volume.} In this step, we define the aliasing in reconstructed LF as a kind of cost, which is similar to the matching cost in the depth estimation task or DIBR task. We term the cost as reconstruction cost. After the reconstruction and the backward shearing, we feed the sheared volume $\{\tilde{L}_{d}(x,y,s)\}_{d\in\mathbb{D}}$ into another neural network, which is termed DIBR network, as shown in Fig.~\ref{fig:CNN}(b). The DIBR network outputs a reconstruction cost volume $\{\mathcal{C}_{d}(x,y,s)\}_{d\in\mathbb{D}}$ that indicates the aliasing degree of the reconstructed LFs. This step can be described as
$$\mathcal{C}_d(x,y,s)=f_{D}(\tilde{L}_d(x,y,s);\vartheta),d\in\mathbb{D},$$
where $f_D(\cdot;\vartheta)$ denotes the DIBR network with trainable parameters $\vartheta$. Detailed configuration of the DIBR network will be introduced in Sec. \ref{Sec:network}. Despite the objective of the proposed Geo-NI being LF reconstruction, we can conveniently render a depth map from the DIBR network. Please refer to Sec. \ref{Sec:depth_render} for detail.

\textbf{Blending.} The final rendered high-angular resolution LF $\hat{L}(x,y,s)$ is obtained by blending the reconstructed LFs sheared by all the depth hypotheses $\{\tilde{L}_{d}(x,y,s)\}_{d\in\mathbb{D}}$. In this perspective, the reconstruction cost volume $\{\mathcal{C}_{d}(x,y,s)\}_{d\in\mathbb{D}}$ is considered as weight maps. While directly weighting each reconstructed LF $\tilde{L}_d$ by reconstruction cost $\mathcal{C}_d$ will cause numerical instability. Therefore, we use normalized cost to formulate the blending
\begin{equation}\label{eq:output}
\begin{split}
\hat{L}(x,y,s)&=\sum_d \tilde{L}_d(x,y,s)\odot\sigma_d(-\mathcal{C}_d(x,y,s)),\\
\sigma_d(-\mathcal{C}_d(x,y,s)) &= \frac{\exp(-\mathcal{C}_d(x,y,s))}{\sum_d\exp(-\mathcal{C}_d(x,y,s))},
\end{split}
\end{equation}
where $\odot$ is the Hadamard product, i.e., element-wise multiplication, and $\sigma_d$ is the softmax function. It should be noted that a smaller reconstruction cost indicates less degree of aliasing effect in the reconstructed LF, i.e., a higher reconstruction quality. We therefore normalize the reconstruction cost volume $\{\mathcal{C}_{d}(x,y,s)\}_{d\in\mathbb{D}}$ using softmin nonlinearity in Eqn.~\ref{eq:output}. Fig. \ref{fig:fusionshear}(b) demonstrates an example of the final rendered high-angular resolution LF (EPI). Since the DIBR part explicitly shows which depth hypothesis is chosen for each pixel, the behaviour of the network is interpretable.

\subsection{Packing-Unpacking Network}\label{Sec:network}
Typical learning-based reconstruction methods~\cite{Eilertsen2017HDR,zhou2018stereo,mildenhall2019local,2020CrossNet} pursue a larger size of receptive field through spatial downsampling to achieve a higher reconstruction quality, e.g., strided convolutions, max-pooling and average-pooling. However, recent researches~\cite{zhang2019making,guizilini20203D,zheng2021rethinking} show that the commonly used downsampling methods ignore the sampling theorem, leading to performance degradation for tasks requiring fine-grained details. In comparison, we propose a novel hierarchical packing-unpacking structure that is able to efficiently increase the receptive field while also preserving high-frequency details for LF reconstruction, as shown in Fig. \ref{fig:pixelshuffle}(a). In the following, we first introduce the proposed packing-unpacking structure and then present the NI and DIBR networks constructed based on the structure.

\subsubsection{Hierarchical packing-unpacking structure}
In a packing block, we encode high spatial resolution LF (features) into high-level features by using a Spatial-to-Channel Pixel Shuffling (S2C-PS)~\cite{shi2016real} followed by a residual module~\cite{he2016deep} of two convolutional layers, as shown in the left part of Fig. \ref{fig:pixelshuffle}(b). Analogously, in an unpacking block, we replace the S2C-PS with its reverse operation, Channel-to-Spatial Pixel Shuffling (C2S-PS), to achieve the decode, as shown in the right part of Fig. \ref{fig:pixelshuffle}(b). Different from the vanilla version in~\cite{shi2016real}, we only fold one spatial dimension of feature maps into extra feature channels since the depth information is mainly weaved in one spatial dimension on the epipolar plane, as shown in Fig. \ref{fig:pixelshuffle}(c). For example, the 5D feature tensor $\phi\in\mathbb{R}^{B\times W\times H\times A\times C}$ (batch, width, height, angular, and channel) is converted to $\phi'\in\mathbb{R}^{B\times W/2\times H\times A\times 2C}$ using the S2C-PS operation.

By stacking the blocks symmetrically, we then construct a hierarchical packing-unpacking structure that reduces the spatial resolution of the input LF in the bottleneck. Since the S2C-PS and S2C-PS are two reversible operations, the proposed packing-unpacking structure is able to gain a large receptive field without high-frequency loss.

\subsubsection{Network architectures}
The networks in our proposed framework can be split into three parts: an encoder that extracts high-level but low-resolution features from high (spatial) resolution input, a bottleneck that performs non-linear mapping, and a decoder that restores high-resolution target, i.e., an LF or a slice of the reconstruction cost volume, from high-level low-resolution features. Table \ref{table:prnet} lists the detailed configurations of the NI and DIBR networks. The encoders and decoders of the NI and DIBR networks incorporate two packing-unpacking blocks symmetrically. We use folding factor of 2 for each packing block and a fold factor of $1/2$ for each unpacking block. In the tables, we term the folding factor as stride factor for convenience. For each residual module in the packing/unpacking block, we apply leaky ReLU activation after the first convolutional layers.

\begin{table}
\caption{Detail configuration of the proposed NI network.}
\vspace{-4mm}
\label{table:prnet}
\begin{center}
\begin{tabular}{l|ccc}
\toprule
Layer & $k$ & $s$ & Output Dim.\\
\midrule
\multicolumn{4}{c}{Encoder}\\
\midrule
Conv1 & [3, 1, 3]  & [1, 1, 1] & $W\times H\times A_1\times 32$\\
Pack.1 & -  & [2, 1, 1] & $\frac{W}{2}\times H\times A_1\times 64$\\
Pack.2 & -  & [2, 1, 1] & $\frac{W}{4}\times H\times A_1\times 128$\\
\midrule
\multicolumn{4}{c}{Residual blocks: $\times 1$ for  NI network, $\times K$ for DIBR network}\\
\midrule
Conv2 & [3, 1, 3]  & [1, 1, 1] & $\frac{W}{4}\times H\times A_1\times 128$\\
Conv3 & [3, 3, 1]  & [1, 1, 1] & $\frac{W}{4}\times H\times A_1\times 128$\\
Conv4 & [3, 1, 3]  & [1, 1, 1] & $\frac{W}{4}\times H\times A_1\times 128$\\
Conv4$\oplus$Pack.2 & -  & - & $\frac{W}{4}\times H\times A_1\times 128$\\
\midrule
\multicolumn{4}{c}{Reconstruction: $\times 1$ for NI network, $\times 0$ for DIBR network}\\
\midrule
DeConv & [5, 1, $2\alpha+1$]  & [1, 1, $\frac{1}{\alpha}$] & $\frac{W}{4}\times H\times A_2\times 128$\\
\midrule
\multicolumn{4}{c}{Decoder}\\
\midrule
Unpack.1 & -  & [$\frac{1}{2}$, 1, 1] & $\frac{W}{2}\times H\times A_2\times 64$\\
Unpack.2 & -  & [$\frac{1}{2}$, 1, 1] & $W\times H\times A_2\times 32$\\
Conv5 & [3, 1, 3]  & [1, 1, 1] & $W\times H\times A_2\times 1$\\
\bottomrule
\end{tabular}
\end{center}
\textit{$k$ denotes the kernel size, $s$ the stride, Conv the 3D convolution layer, Deconv the 3D deconvolution layer and $\oplus$ the element-wise addition. We ignore the batch dimension of tensors here.}
\vspace{-4mm}
\end{table}

The main difference between the NI and DIBR networks is as follows. For the NI network, we use one residual block ($K=1$) in the bottleneck. We insert a deconvolutional layer (also known as transposed convolution layer) of stride [1, 1, $1/\alpha$] at the end of the bottleneck to achieve plenoptic reconstruction. Thus, the angular resolutions of the output features are $A_1=A$ and $A_2=\alpha(A-1)+1$. For the DIBR network, we stack $K=3$ residual blocks in the bottleneck. The angular resolutions of the output features are $A_1=A_2=\alpha(A-1)+1$.

In our practical implementation, we do not employ skip connections (like those in standard U-net~\cite{Olaf2015unet}) to concatenate low-level features in the encoder with high-level features in the decoder. Because the residual block~\cite{he2016deep} has already provided a path for the flow of information and gradients throughout the network. Please refer to Sec. \ref{Sec:ablation} for the comparison between the proposed packing-unpacking structure and standard U-net.

\subsubsection{Modeling 4D cost volume with 3D convolution}
The input and output (reconstruction cost volume) of the networks are 4-dimensional tensors of shape $|\mathbb{D}|\times W\times H\times A$ (shear, width, height, and angular) neglecting the batch and channel dimensions. A straightforward implementation is to employ a network that is fully convolutional along these dimensions, which inevitably leading to an extremely large number of network parameters due to the requirement of 4D convolution. 

Fortunately, each slice $\mathcal{H}_d(L)$ in the reconstruction task (NI network) or $\tilde{L}_d$ in the cost assignment task (DIBR network) is independent along the shear dimension. Therefore, we can feed them into the networks slice by slice, i.e., $\mathcal{H}_d(L)$ or $\tilde{L}_d$. In practice, we employ a more efficient implementation by folding the shear dimension of the input into the batch dimension ($B|\mathbb{D}|\times W\times H\times A\times C$), and unfolding it after the inference ($B\times \mathbb{D}\times W\times H\times A\times C$). Then the reconstructed light field volume $\{\bar{L}_{d}\}_{d\in\mathbb{D}}$ or the reconstruction cost volume $\{\mathcal{C}_d\}_{d\in\mathbb{D}}$ can be obtained with a single forward propagation. This modeling strategy also brings an important feature that the proposed Geo-NI is able to apply a flexible settings of depth hypothesis without retraining, please refer to Sec.~\ref{Sec:ablation_shear} for details.

\subsection{Implementation}\label{Sec:training}
\subsubsection{Training objective}
In a narrow sense, the objectives between the NI and DIBR networks in our framework are different. The objective of the NI network is to reconstruct a high angular resolution LF. While the objective of the DIBR network is to evaluate whether the LF is well reconstructed under a certain depth hypothesis. The latter objective is hard to achieve when the ground truth depth is unavailable. Fortunately, from a macro perspective, the objective of the entire framework is to synthesize a high-quality LF. In addition, all modules in our framework are differentiable, and thus, the proposed Geo-NI framework can be trained in an end-to-end manner. 

\begin{figure}
\begin{center}
\includegraphics[width=1\linewidth]{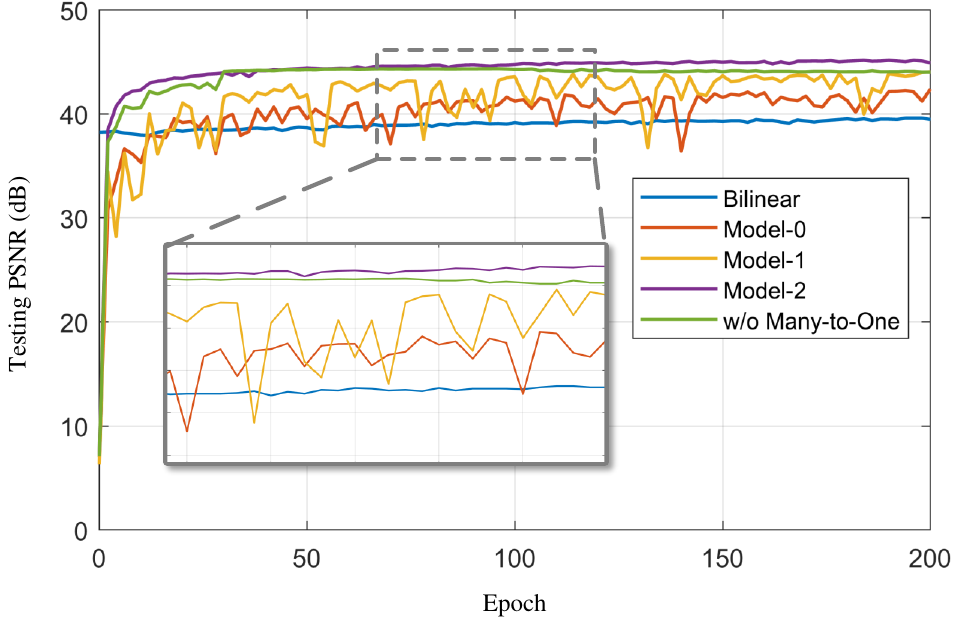}
\end{center}
\vspace{-4mm}
   \caption{We apply a Many-to-One strategy to optimize the DIBR network using randomly activated NI networks. The figure demonstrate the convergence curves (testing PSNR) of the multiple NI networks (including a bilinear interpolation) and the curve without using the Many-to-One strategy.}
\label{fig:loss}
\vspace{-3mm}
\end{figure}

We measure the $L_1$ distance between the reconstructed light field $\tilde{L}$ and the desired high angular resolution LF $L$ to optimize the parameters $\theta$ in the NI network, and the $L_1$ distance between the final output $\hat{L}$ and the desired LF $L$ to optimize the parameters $\vartheta$ in the DIBR network. The overall training objective is formulated as follows
$$
\min_{\theta,\vartheta}\sum_n\Vert\tilde{L}^{(n)}_{d^*}-L^{(n)}\Vert_1+\Vert\hat{L}^{(n)}\odot\mathcal{M}-L^{(n)}\odot\mathcal{M}\Vert_1,
$$
where $\tilde{L}_{d^*}$ indicates the LF with shear amount $d=0$, $n$ denotes the $n^{\text{th}}$ training instance, $\mathcal{M}$ is a binary mask that avoids computing the loss on pixels that do not have valid values caused by the shearing operation, and $\odot$ denotes element-wise multiplication.

\begin{figure*}
\begin{center}
\includegraphics[width=1\linewidth]{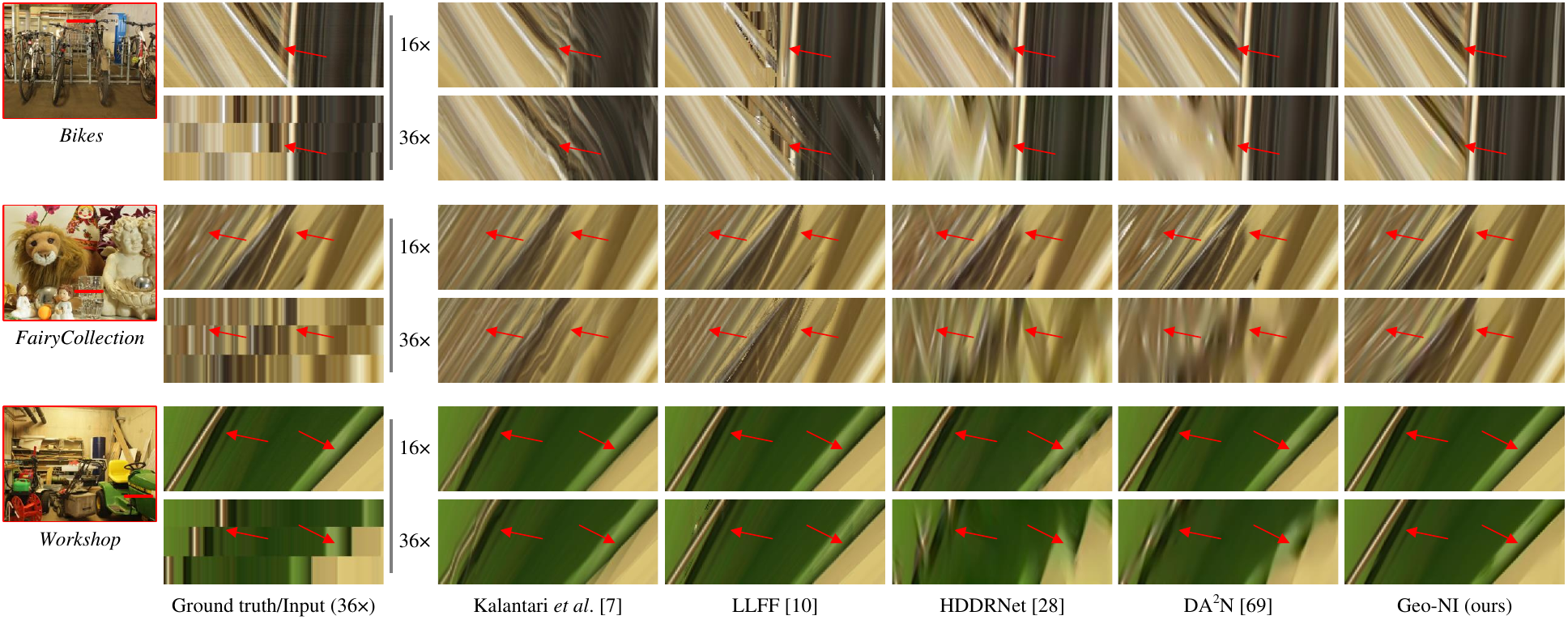}
\end{center}
\vspace{-4mm}
   \caption{Comparison of the results (reconstructed EPIs) on LFs from the MPI Light Field Archive~\cite{kiran2017towards} (reconstruction scales $16\times$ and $36\times$). The second column shows the ground truth EPI and the input EPI for $36\times$ reconstruction.}
\vspace{-2mm}
\label{fig:Result1}
\end{figure*}
                                                                                                                                                                                                                                                                                                                                                                                                                                                                           
\begin{table*}
\caption{Quantitative results (PSNR/SSIM) of reconstructed light fields on LFs from the MPI Light Field Archive~\cite{kiran2017towards}.}
\label{table:Result1}
\vspace{-3mm}
\begin{center}
\begin{tabular}{p{2.4cm}|c|p{1.9cm}<{\centering} p{1.9cm}<{\centering} p{1.9cm}<{\centering} p{1.9cm}<{\centering} p{1.9cm}<{\centering} p{1.9cm}<{\centering}}
\toprule
& Scale & \textit{Bikes} & \textit{FairyCollection} & \textit{LivingRoom} & \textit{Mannequin} & \textit{WorkShop} & Average\\
\midrule
Kalantari~\textit{et al.}~\cite{DoubleCNN}& \multirow{7}*{$16\times$} & 30.67 / 0.935 & 32.39 / 0.952 & 41.62 / 0.973 & 37.15 / 0.970  & 33.94 / 0.971 & 35.15 / 0.960\\
Wu~\textit{et al.}~\cite{WuEPICNN2018}& & 31.22 / 0.951 & 30.33 / 0.942 & 42.43 / 0.991 & 39.53 / 0.989 & 33.49 / 0.977 & 35.40 / 0.970\\
Yeung~\textit{et al.}~\cite{YeungECCV2018} & & 32.67 / 0.967 & 31.82 / 0.969 & 43.54 / 0.993 & 40.82 / 0.992 & 37.21 / 0.988 & 37.21 / 0.982\\
LLFF~\cite{mildenhall2019local} & & 34.95 / 0.963 & 34.01 / 0.966 & 44.73 / 0.987 & 39.92 / 0.985 & 37.61 / 0.985 & 38.24 / 0.977\\
HDDRNet~\cite{meng2019high} & & 33.97 / 0.976 & 35.08 / 0.979 & 44.83 / 0.997 & 40.60 / 0.993 & 38.54 / 0.992 & 38.60 / 0.987\\
DA$^2$N~\cite{wu2021revisiting} & & 35.79 / 0.984 & 36.23 / 0.981 & 45.91 / 0.996 & 40.83 / 0.992 & 40.11 / 0.994 & 39.77 / 0.990\\
Geo-NI (ours) & &\textbf{37.02} / \textbf{0.989} & \textbf{38.84} / \textbf{0.991} & \textbf{46.87} / \textbf{0.997} & \textbf{41.57} / \textbf{0.993}  & \textbf{41.80} / \textbf{0.996} & \textbf{41.22} / \textbf{0.993}\\
\midrule
Kalantari~\textit{et al.}~\cite{DoubleCNN}& \multirow{11}*{$36\times$} & 26.99 / 0.869 & 28.34 / 0.905 & 37.10 / 0.929 & 33.82 / 0.945  & 29.61 / 0.938 & 31.71 / 0.917\\
Wu~\textit{et al.}~\cite{WuEPICNN2018}& & 25.44 / 0.856 & 23.60 / 0.807 & 35.31 / 0.948 & 31.79 / 0.945 & 25.42 / 0.873 & 28.31 / 0.886\\
Yeung~\textit{et al.}~\cite{YeungECCV2018} & & 26.92 / 0.896 &  27.12 / 0.897 & 37.44 / 0.970 &  33.77 / 0.963 &  28.70 / 0.932 & 30.79 / 0.932\\
LLFF~\cite{mildenhall2019local} & & 27.40 / 0.892 & 28.56 / 0.918 & 39.54 / 0.980 & 33.53 / 0.949 & 30.12 / 0.949 & 31.83 / 0.938\\
HDDRNet~\cite{meng2019high} & &  26.35 / 0.886 & 24.50 / 0.853 & 36.17 / 0.966 & 32.47 / 0.960 & 27.16 / 0.919 & 29.33 / 0.917\\
DA$^2$N~\cite{wu2021revisiting} & &  27.94 / 0.917 & 26.52 / 0.903 & 38.39 / 0.975 & 35.70 / 0.976 & 29.80 / 0.961 & 30.67 / 0.932\\
MVSNeRF~\cite{Chen2021MVSNeRFFG} & &  26.34 / 0.864 & 25.57 / 0.836 & 33.67 / 0.916 & 30.07 / 0.907 & 24.73 / 0.886 & 28.08 / 0.882\\
MVSNeRF-FT~\cite{Chen2021MVSNeRFFG} & & 27.76 / 0.905 & 28.70 / 0.903 & 35.85 / 0.931 & 30.55 / 0.914 & 26.08 / 0.901 & 29.79 / 0.911\\
IBRNet~\cite{Wang_2021_CVPR} & &  27.11 / 0.919 & 30.14 / 0.957 & 38.57 / 0.983 & 33.24 / 0.963 & 31.57 / 0.970 & 32.12 / 0.958\\
IBRNet-FT~\cite{Wang_2021_CVPR} & &  29.84 / 0.947 & 31.33 / 0.962 & 39.75 / 0.987 & 34.81 / 0.973 & 33.12 / 0.977 & 33.77 / 0.969\\
Geo-NI (ours) & & \textbf{31.41} / \textbf{0.963} & \textbf{32.62} / \textbf{0.970} & \textbf{43.35} / \textbf{0.994} & \textbf{36.65} / \textbf{0.981} & \textbf{34.96} / \textbf{0.985} & \textbf{35.80} / \textbf{0.979}\\
\bottomrule
\end{tabular}
\end{center}
\vspace{-5mm}
\end{table*}

\subsubsection{Training Data}\label{Sec:train_data}
We train the networks in the proposed Geo-NI framework by using LFs from the Stanford (New) Light Field Archive~\cite{StanfordLFdatasets}, which contains 12 LFs with $17\times17$ views (the \textit{Lego Gantry Self Portrait} is excluded due to the moving object in the scene). Since the network input is 3D LFs, we can extract 17 $L(x,y,s)$ and 17 $L(y,x,t)$ in each 4D LF set. To enhance the performance of the DIBR network, we augment the extracted 3D LFs using the shearing operation in Eqn.~\ref{eq:shearing} with shear amounts $d=\pm2$. This augmentation increases the number of training examples by 2 times. In the training procedure, we crop the extracted 3D LFs into sub-LFs with spatial resolution $128\times18$ (width and height) and a stride of 40 pixels. We have three settings with respect to the reconstruction factors $\alpha=4,7$ in the NI network. Thus, the input/output angular resolution of the training samples for the Geo-NI framework are 5/17 and 3/15, respectively. Although the reconstruction factor of the network is fixed, we can achieve a flexible upsampling rate through network cascade.

\subsubsection{Training Details}\label{Sec:train_details}
To make sure the DIBR network can detect the aliasing effects instead of remembering certain patterns in the reconstructed LF slices, we develop a Many-to-One strategy during the training phase. More specifically, we randomly activate one NI network among multiple candidates to optimize the DIBR network. The candidates include models with different numbers of packing-unpacking structures (i.e., different sizes of receptive field) as well as a simple bilinear interpolation. Note that the bilinear interpolation has no trainable parameter and can only perform aliasing-free reconstruction for disparity within 1 pixel. Fig. \ref{fig:loss} demonstrates the convergence curves (average PSNR on the test data) of the multiple candidate NI networks in the training phase, where Bilinear stands for the bilinear interpolation, Model-0 stands for the NI network without packing-unpacking structure and Model-1 (Model-2) stands for the NI network with one (two) packing-unpacking structure(s). The result indicates that without using the proposed Many-to-One strategy, the performance of the Geo-NI will degrade along with the training procedure, suggesting overfitting of the DIBR network.

\begin{figure*}
\begin{center}
\includegraphics[width=1\linewidth]{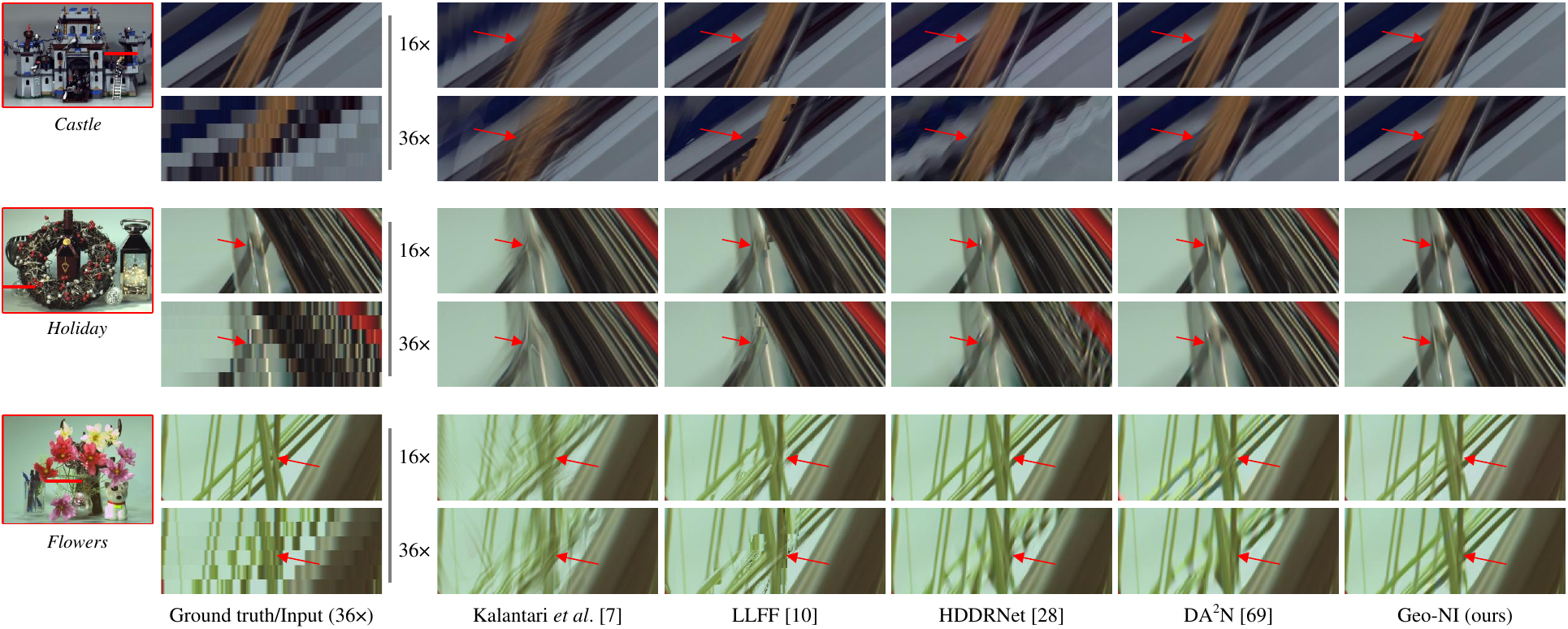}
\end{center}
\vspace{-4mm}
   \caption{Comparison of the results on LFs from the CIVIT Dataset~\cite{ICME2018} (reconstruction scales $16\times$ and $36\times$). The second column shows the ground truth EPI and the input EPI for $36\times$ reconstruction.}
\label{fig:Result2}
\vspace{-3mm}
\end{figure*}
                                                                                                                                                                                                                                                                                                                                                                                                                                                                           
\begin{table*}
\caption{Quantitative results (PSNR/SSIM) of reconstructed light fields on LFs from the CIVIT Dataset~\cite{ICME2018}.}
\label{table:Result2}
\vspace{-4mm}
\begin{center}
\begin{tabular}{p{2.5cm}|c|p{1.9cm}<{\centering} p{1.9cm}<{\centering} p{1.9cm}<{\centering} p{1.9cm}<{\centering} p{1.9cm}<{\centering} p{1.9cm}<{\centering}}
\toprule
& Scale & \textit{Seal \& Balls} & \textit{Castle} & \textit{Holiday} & \textit{Dragon} & \textit{Flowers} & Average\\
\midrule
Kalantari~\textit{et al.}~\cite{DoubleCNN}& \multirow{7}*{$16\times$} & 43.13 / 0.985 & 36.03 / 0.965 & 32.44 / 0.961 & 39.50 / 0.985 & 35.21 / 0.973 & 37.26 / 0.974\\
Wu~\textit{et al.}~\cite{WuEPICNN2018}& & 45.21 / 0.994 & 35.20 / 0.977 & 35.58 / 0.987 & 46.39 / 0.997 & 41.60 / 0.995 & 40.80 / 0.990\\
Yeung~\textit{et al.}~\cite{YeungECCV2018} & & 44.38 / 0.992 & 37.86 / 0.989 & 36.06 / 0.988 & 45.52 / 0.997 & 42.30 / 0.994 & 41.22 / 0.992\\
LLFF~\cite{mildenhall2019local} & & 45.50 / 0.990 & 38.60 / 0.971 & 36.69 / 0.984 & 44.80 / 0.992 & 41.19 / 0.989 & 41.36 / 0.985\\
HDDRNet~\cite{meng2019high}&& 44.24 / 0.997 & 39.88 / 0.991 & 38.09 / 0.992 & 44.26 / 0.997 & 42.04 / 0.996 & 41.70 / 0.995\\
DA$^2$N~\cite{wu2021revisiting}&& 46.19 / 0.996 & 40.77 / 0.992 & 37.99 / 0.992 & 47.19 / 0.998 & 41.95 / 0.996 & 42.82 / 0.995\\
Geo-NI (ours) & & \textbf{48.82} / \textbf{0.998} & \textbf{41.40} / \textbf{0.992} & \textbf{38.99} / \textbf{0.993} & \textbf{48.16} / \textbf{0.998}  & \textbf{44.23} / \textbf{0.997} & \textbf{44.32} / \textbf{0.996}\\
\midrule
Kalantari~\textit{et al.}~\cite{DoubleCNN}& \multirow{11}*{$36\times$} &  38.01 / 0.977 & 32.95 / 0.948 & 29.11 / 0.928 & 35.49 / 0.975 & 32.51 / 0.959 & 33.61 / 0.957\\
Wu~\textit{et al.}~\cite{WuEPICNN2018}& & 37.34 / 0.969 & 31.15 / 0.960 & 27.99 / 0.927 &  33.77 / 0.974 &  34.02 / 0.977 & 32.54 / 0.961\\
Yeung~\textit{et al.}~\cite{YeungECCV2018} & & 38.56 / 0.979 &  33.12 / 0.971 &  29.97 / 0.952 &  36.95 / 0.986 & 34.93 / 0.982 & 34.71 / 0.974\\
LLFF~\cite{mildenhall2019local} & & 40.55 / 0.982 & 33.95 / 0.954 & 30.16 / 0.941 & 37.99 / 0.980 & 33.50 / 0.969 & 35.23 / 0.965\\
HDDRNet~\cite{meng2019high}&& 40.15 / 0.984 &  33.35 / 0.972 & 30.62 / 0.957 &  35.83 / 0.985 &  36.76 / 0.988 & 35.34 / 0.977\\
DA$^2$N~\cite{wu2021revisiting}&& 43.96 / 0.992 &  36.58 / 0.983 &  32.78 / 0.973 &  43.61 / 0.996 &  36.67 / 0.988 & 38.72 / 0.986\\
MVSNeRF~\cite{Chen2021MVSNeRFFG} & &  28.41 / 0.932 & 27.90 / 0.914 & 25.38 / 0.904 & 29.47 / 0.910 & 30.63 / 0.904 & 28.36 / 0.913\\
MVSNeRF-FT~\cite{Chen2021MVSNeRFFG} & &  31.31 / 0.943 & 29.99 / 0.927 & 26.30 / 0.914 & 30.11 / 0.929 & 33.38 / 0.933 & 30.22 / 0.929\\
IBRNet~\cite{Wang_2021_CVPR} & &  40.90 / 0.990 &  34.14 / 0.979 & 29.69 / 0.955 & 38.79 / 0.993 & 34.37 / 0.982 & 35.58 / 0.980\\
IBRNet-FT~\cite{Wang_2021_CVPR} & & 42.75 / 0.993 & 35.27 / 0.981 & 31.54 / 0.968 & 42.31 / 0.996 & 35.83 / 0.986 & 37.54 / 0.985 \\
Geo-NI (ours) & & \textbf{45.40} / \textbf{0.996} & \textbf{37.85} / \textbf{0.987} & \textbf{33.62} / \textbf{0.978} & \textbf{45.10} / \textbf{0.997} & \textbf{39.18} / \textbf{0.993} & \textbf{40.23} / \textbf{0.990}\\
\bottomrule
\end{tabular}
\end{center}
\vspace{-4mm}
\end{table*}

In our implementation, the training is performed on the luminance channel (Y channel) in the YCbCr color space. The parameters in the networks are initialized with values drawn from the normal distribution (zero mean and standard deviation of $1\times10^{-2}$). We optimize the networks by using ADAM solver~\cite{Kingma2014Adam} with a learning rate of $1\times10^{-4}$ ($\beta_1=0.9$, $\beta_2=0.999$) and a mini-batch size of 8. The network converges after $200$ epochs on the training dataset. The proposed Geo-NI is implemented in the \emph{Pytorch} framework~\cite{Pytorch}.

\section{Evaluations}
In this section, we evaluate the proposed Geo-NI framework on various kinds of LFs, including those from both gantry systems and plenoptic cameras (Lytro Illum~\cite{Lytro}). We mainly compare our method with two depth-based methods, Kalantari~\textit{et al.}~\cite{DoubleCNN} (depth-based) and LLFF~\cite{mildenhall2019local} (MPI representation) and four depth-independent methods, Wu~\textit{et al.}~\cite{WuEPICNN2018}, Yeung~\textit{et al.}~\cite{YeungECCV2018}, HDDRNet~\cite{meng2019high} and DA$^2$N~\cite{wu2021revisiting}. The quantitative evaluations are reported by measuring the average PSNR and SSIM~\cite{SSIM} values over the synthesized views of the luminance channel in the YCbCr space. Please refer to the submitted video for more visual results.

\subsection{Evaluations on Light Fields from Gantry Systems}\label{sec:gantry_exp}
\subsubsection{Experiment settings}
The comparisons are performed on 3D LFs from the MPI Light Field Archive~\cite{kiran2017towards} and the CIVIT Dataset~\cite{ICME2018}. The resolutions of the datasets are $960\times720\times101$ and $1280\times720\times193$ (width, height and angular). In this experiment, we use two upsampling scale settings to evaluate the capability of the proposed Geo-NI: $16\times$ reconstruction using 7/13 views as input (MPI Light Field Archive~\cite{kiran2017towards}/CIVIT Dataset~\cite{ICME2018}) and $36\times$ reconstruction using 3/6 views as input. The depth hypotheses $d\in\mathbb{D}$ for the shearing operation are set to $\{0,\pm4,\pm8\}$ for $16\times$ reconstruction and $\{0,\pm4,\pm8,\pm12,\pm16,\pm20,\pm24\}$ for $36\times$ reconstruction, respectively.

\subsubsection{Qualitative comparison}
Fig. \ref{fig:Result1} shows three reconstruction results on LFs (\textit{Bikes}, \textit{FairyCollection} and \textit{WorkShop}) from the MPI Light Field Archive~\cite{kiran2017towards}. And Fig. \ref{fig:Result2} shows the reconstruction results on LFs (\textit{Castle}, \textit{Holiday}, and \textit{Flowers}) from the CIVIT Dataset~\cite{ICME2018}. We demonstrate results with a reconstruction scale $16\times$ as well as $36\times$ on these two datasets. The maximum disparity range of the $36\times$ reconstruction reaches 75 pixels in the \textit{FairyCollection} case~\cite{kiran2017towards}.

\begin{figure}
\begin{center}
\includegraphics[width=1\linewidth]{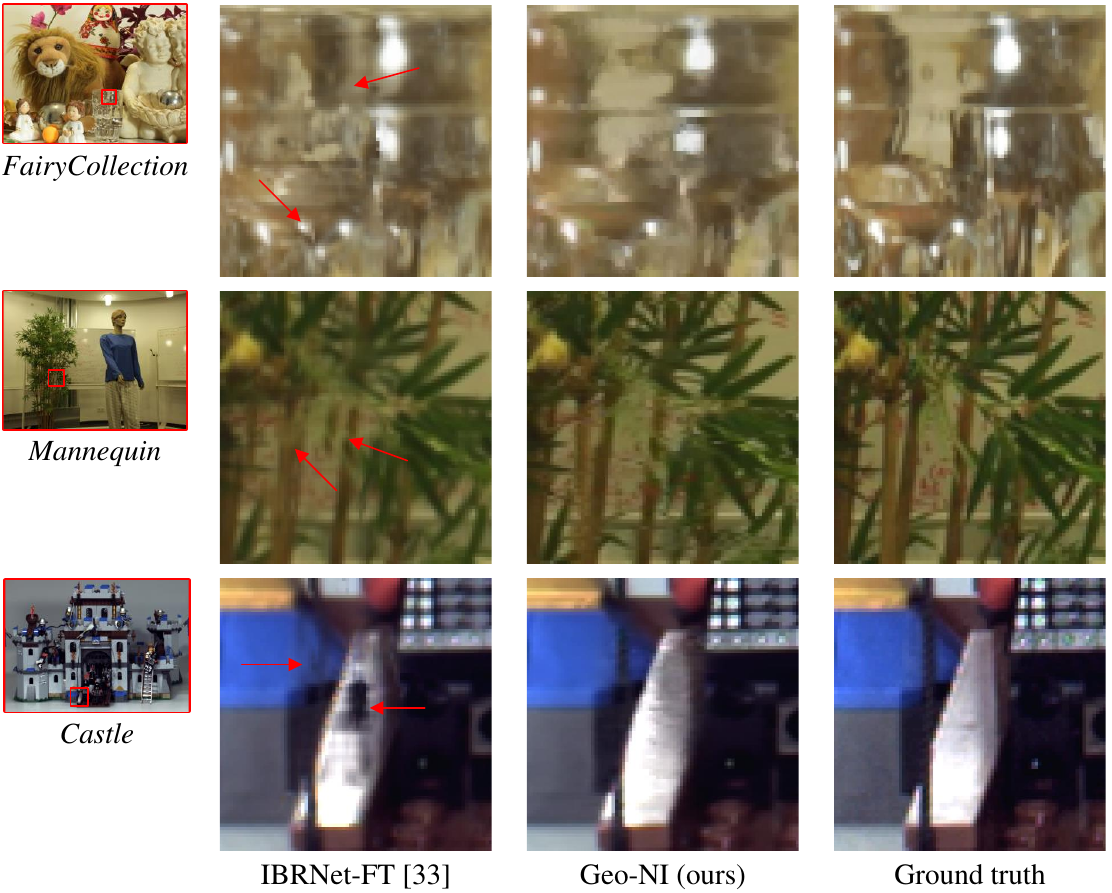}
\end{center}
\vspace{-4mm}
   \caption{Comparison with a state-of-the-art NeRF-based method, IBRNet~\cite{Wang_2021_CVPR} (reconstruction scale $36\times$). The LFs are from the MPI Light Field Archive~\cite{kiran2017towards} and the CIVIT Dataset~\cite{ICME2018}.}
\label{fig:nerf}
\vspace{-3mm}
\end{figure}

The qualitative comparisons indicate that the proposed Geo-NI framework is able to produce high-quality reconstruction on scenes with large disparities or non-Lambertian effects. The depth-based methods, Kalantari~\textit{et al.}~\cite{DoubleCNN} and LLFF~\cite{mildenhall2019local}, show high performance on simple structured scenes with large disparities while the depth-independent methods, HDDRNet~\cite{meng2019high} and DA$^2$N~\cite{wu2021revisiting}, appear aliasing effects especially for $36\times$ reconstruction, e.g., the \textit{WorkShop} case in Fig. \ref{fig:Result1}. For scenes with complex structures or non-Lambertian effects but with relatively small disparities, the depth-independent methods, HDDRNet~\cite{meng2019high} and DA$^2$N~\cite{wu2021revisiting}, outperforms the depth-based methods, Kalantari~\textit{et al.}~\cite{DoubleCNN} and LLFF~\cite{mildenhall2019local}, e.g., the cases \textit{Bikes} and \textit{FairyCollection} in Fig. \ref{fig:Result1} and the three cases in Fig. \ref{fig:Result2}. However, the original structure or non-Lambertian effects will not be preserved in the results by the baseline methods when the input views are extremely sparse ($36\times$). The proposed Geo-NI is able to effectively combine the advantages of depth-based methods and depth-independent methods, and produces plausible reconstruction results on challenge cases with large disparities, complex structures or non-Lambertian effects.

Since the vanilla version of the network by Yeung~\textit{et al.}~\cite{YeungECCV2018} and Meng \textit{et al.}~\cite{meng2019high} (HDDRNet) were specifically designed for 4D LFs, we modify their convolutional layers to fit the 3D input while keeping its network architecture unchanged. We re-train the networks in the state-of-the-art methods (Kalantari~\textit{et al.}~\cite{DoubleCNN}, Yeung~\textit{et al.}~\cite{YeungECCV2018}, LLFF~\cite{mildenhall2019local} and HDDRNet~\cite{meng2019high} by using the same training dataset as the proposed Geo-NI. Due to the particularity of the training datasets, we compare DA$^2$N~\cite{wu2021revisiting} using the released network parameters. We perform network cascade to achieve different upsampling scales, i.e., two (three) cascades for $16\times$ ($36\times$) upsampling using a network of reconstruction factor $\alpha=4$.


In addition, we compare the proposed method with two state-of-the-art Neural Radiance Fields (NeRF)-based methods, MVSNeRF~\cite{Chen2021MVSNeRFFG} and IBRNet~\cite{Wang_2021_CVPR} at the $36\times$ reconstruction scale. We also fine-tune the pre-trained MVSNeRF and IBRNet models on the non-test views for each test scene. For example, for the $36\times$ reconstruction task on the MPI Light Field Archive~\cite{kiran2017towards}, the views from \#74 to \#101 are used for the fine-tuning. Fig. \ref{fig:nerf} shows the visual comparison with the fine-tuned IBRNet~\cite{Wang_2021_CVPR} (IBRNet-FT). Although the IBRNet employs a per-scene fine-tuning procedure, the proposed Geo-NI produces more pleasant results with challenging refractive effects (the first case), complex occlusions (the second case), or the specular effects (the third case).

\begin{figure*}
	\begin{center}
		\includegraphics[width=1\linewidth]{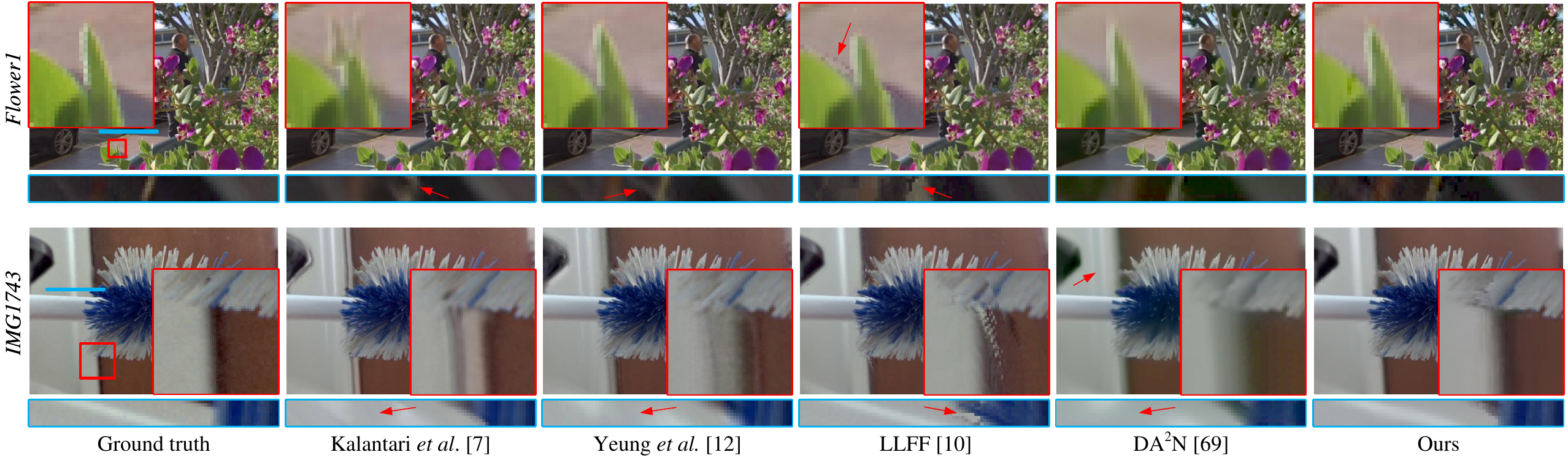}
	\end{center}
	\vspace{-4mm}
	\caption{Comparison of the results on LFs from Lytro Illum. The results show the reconstructed views, zoom-in regions and the EPIs at the location marked by red lines. LFs are \textit{Flowers1} and \textit{IMG1743} from the \textit{30 Scenes}~\cite{DoubleCNN}.}
	\label{fig:Result3}
\end{figure*}

\subsubsection{Quantitative comparison}
The quantitative results of $16\times$ and $36\times$ reconstructions for the MPI Light Field Archive~\cite{kiran2017towards} and the CIVIT Dataset~\cite{ICME2018} are shown in Table \ref{table:Result1} and Table \ref{table:Result2}, respectively. We also evaluate the results of the state-of-the-art NeRF-based methods, including the pretrained models, MVSNeRF~\cite{Chen2021MVSNeRFFG} and IBRNet~\cite{Wang_2021_CVPR}, and there per-scene fine-tuned versions, MVSNeRF-FT and IBRNet-FT, at the $36\times$ reconstruction scale. For LFs with relatively small disparities ($16\times$ reconstruction), the methods without any depth information by Wu~\textit{et al.}~\cite{WuEPICNN2018}, Yeung~\textit{et al.}~\cite{YeungECCV2018} and Meng~\textit{et al.}~\cite{meng2019high} (HDDRNet) is comparable to the depth-based methods by Kalantari~\textit{et al.}~\cite{DoubleCNN} and Mildenhall~\textit{et al.}~\cite{mildenhall2019local} (LLFF). However, the performances of the depth-free methods drop quickly when the disparity is large. For example, on the CIVIT Dataset~\cite{ICME2018} (Table~\ref{table:Result2}), the PSNR of the method by Wu~\textit{et al.}~\cite{WuEPICNN2018} is 3dB higher than the method by Kalantari~\textit{et al.}~\cite{DoubleCNN} for $16\times$ reconstruction but 1dB lower for $36\times$ reconstruction. The NeRF-based method IBRNet achieves better results than most of the compared methods, especially on scenes with large disparities as shown in Table \ref{table:Result1}. Compared with the baseline methods, the proposed Geo-NI framework shows superior performance on both LF datasets from gantry systems.

\subsection{Evaluations on Light Fields from Lytro Illum}
\subsubsection{Experiment settings}
The comparisons are performed on 4D LFs from three Lytro Illum datasets, the \textit{30 Scenes} dataset (30 LFs) by Kalantari~\textit{et al.}~\cite{DoubleCNN}, and the \textit{Reflective} (32 LFs) and \textit{Occlusions} (51 LFs) categories from the Stanford Lytro Light Field Archive~\cite{StanfordLytro}. In this experiment, the upsampling factor is set to $7\times$ using $2\times2$ views as input to reconstruct an $8\times8$ LF. The depth hypotheses $d\in D$ for the shearing operation is set to $\{0,\pm4,\pm8,\pm12,\pm16\}$.

In addition to the methods evaluated in Sec.~\ref{sec:gantry_exp}, we also compare the proposed Geo-NI framework with a depth-based method by Meng~\textit{et al.}~\cite{meng2021light}. Since the vanilla versions of the networks in~\cite{DoubleCNN,YeungECCV2018,wang2020high,meng2019high,meng2021light} are trained on Lytro LFs, we use their released network parameters without re-training. The networks by Wu~\textit{et al.}~\cite{WuEPICNN2018} and Mildenhall~\textit{et al.}~\cite{mildenhall2019local} (LLFF) are re-trained using the same dataset introduced in Sec. \ref{Sec:train_data}.

\subsubsection{Qualitative comparison}
Fig. \ref{fig:Result3} shows two results for $7\times$ reconstruction ($2\times2\rightarrow8\times8$) on LFs (\textit{Flower1} and \textit{IMG1743}) from the \textit{30 Scenes}~\cite{DoubleCNN}. In the \textit{IMG1743} case, the maximum disparity reaches about 26 pixels between input views.

In the first case, the depth-independent methods by Yeung~\textit{et al.}~\cite{YeungECCV2018} and DA$^2$N~\cite{wu2021revisiting} produce promising results due to the relatively small disparities. While the depth-based methods by Kalantari~\textit{et al.}~\cite{DoubleCNN} and LLFF~\cite{mildenhall2019local} produce ghosting or tearing artifacts around occlusion boundaries. In the second case, the result by Yeung~\textit{et al.}~\cite{YeungECCV2018} appears aliasing effects in the background due to the large disparity. While the DA$^2$N~\cite{wu2021revisiting} network generates blurry results as shown by the background door in the zoom-in figure and EPI. For the depth-based methods, the networks by Kalantari~\textit{et al.}~\cite{DoubleCNN} produce severe aliasing effects due to errors in depth estimation. The MPI-based method, LLFF~\cite{mildenhall2019local}, shows plausible results in the second case. However, the plane assignment error introduces tearing artifacts in the rendered view, as shown by the zoom-in figure. In comparison, the proposed Geo-NI framework provides reconstructed LFs with higher view consistency (as shown in the demonstrated EPIs).

\subsubsection{Quantitative comparison}
Table~\ref{table:Result3} lists the quantitative results (PSNR/SSIM) averaged on LFs in each dataset. Limited by the baseline between viewpoints, the disparity range of LFs from Lytro Illum is smaller than that from the gantry system, despite we only sample two viewpoints in each angular dimension, i.e., generally smaller than 14 pixels. Therefore, the depth-independent methods are able to achieve comparable or even superior performances than the depth-based methods. Since the DIBR network is able to elaborately select LFs with the highest reconstruction quality, the proposed Geo-NI achieves the highest PSNR and SSIM values among the depth-based and depth-independent methods.

\begin{table}
\caption{Quantitative results (PSNR/SSIM) of reconstructed views on LFs from Lytro Illum~\cite{Lytro} ($7\times$ reconstruction, $2\times2 \rightarrow8\times8$).}
\label{table:Result3}
\vspace{-3mm}
\begin{center}
\begin{tabular}{l|ccc}
\toprule
&  \textit{30 Scenes}~\cite{DoubleCNN}  &\textit{Reflective}~\cite{StanfordLytro} &\textit{Occlusions}~\cite{StanfordLytro}\\
\midrule
Kalantari~\textit{et al.}~\cite{DoubleCNN}& 38.21 / 0.974 & 35.84 / 0.942 & 31.81 / 0.895 \\
Wu~\textit{et al.}~\cite{WuEPICNN2018}& 36.28 / 0.965 & 36.48 / 0.962 & 32.19 / 0.907 \\
Yeung~\textit{et al.}~\cite{YeungECCV2018}& 39.22 / 0.977 & 36.47 / 0.947  & 32.68 / 0.906\\
LLFF~\cite{mildenhall2019local} & 38.17 / 0.974 & 36.40 / 0.948 & 31.96 / 0.901 \\
HDDRNet~\cite{meng2019high} & 38.33 / 0.967 & 36.77 / 0.931  & 32.78 / 0.909\\
Meng~\textit{et al.}~\cite{meng2021light} & 39.14 / 0.970 & 37.01 / 0.950  & 33.10 / 0.912\\
DA$^2$N~\cite{wu2021revisiting}   &        38.99 / 0.986 & 36.72 / 0.975 & 33.14 / 0.950 \\
DistgASR~\cite{ying2021disentangling} &    39.19 / 0.990 & 36.02 / 0.967 & 33.67 / 0.954 \\
Geo-NI (ours) & \textbf{40.68} / \textbf{0.990} & \textbf{38.05} / \textbf{0.977} & \textbf{34.54} / \textbf{0.964} \\
\bottomrule
\end{tabular}
\end{center}
\vspace{-4mm}
\end{table}

\subsection{Ablation studies}\label{Sec:ablation}
In this subsection, we empirically investigate the modules in the proposed Geo-NI framework by performing the following ablation studies. 


\begin{table}\scriptsize
\caption{Ablation studies (PSNR/SSIM) of different settings of our proposed modules on LFs from the MPI Light Field Archive~\cite{kiran2017towards} and the CIVIT Dataset~\cite{ICME2018}.}
\label{table:ablations1}
\vspace{-3mm}
\begin{center}
\begin{tabular}{p{0.1cm}|p{0.15cm}<{\centering} p{0.3cm}<{\centering}p{0.4cm}<{\centering} | p{0.45cm}<{\centering} |  p{2.3cm}<{\centering} p{2.1cm}<{\centering}}
\toprule
& NI & DIBR & Pack. & Scale & MPI LF Archive~\cite{kiran2017towards} & CIVIT Dataset~\cite{ICME2018}\\
\midrule
1 & \ding{55} & \checkmark & \checkmark & \multirow{4}*{$16\times$} & 36.50 / 0.987 & 37.17 / 0.993\\
2 & \checkmark & \ding{55} & \checkmark & & 41.03 / 0.992 & \textbf{44.62} / \textbf{0.996}\\
3 & \checkmark & \checkmark & \ding{55} & & 39.90 / 0.990 & 43.44 / 0.995\\
4 & \checkmark & \checkmark & \checkmark & & \textbf{41.22} / \textbf{0.993} & 44.32 / \textbf{0.996}\\
\midrule
1 & \ding{55} & \checkmark & \checkmark & \multirow{4}*{$36\times$} & 32.72 / 0.973 & 35.45 / 0.984 \\
2 & \checkmark & \ding{55} & \checkmark & &  32.66 / 0.954 &  38.98 / 0.987 \\
3 & \checkmark & \checkmark & \ding{55} & & 33.01 / 0.959 & 38.93 / 0.985 \\
4 & \checkmark & \checkmark & \checkmark & & \textbf{35.80} / \textbf{0.979} & \textbf{40.23} / \textbf{0.990} \\
\bottomrule
\end{tabular}
\end{center}
\vspace{-4mm}
\end{table}

\begin{table}\scriptsize
\caption{Ablation studies (PSNR/SSIM) of different settings of our proposed modules on LFs from Lytro Illum~\cite{Lytro} ($7\times$ reconstruction, $2\times2 \rightarrow8\times8$).}
\label{table:ablations2}
\vspace{-3mm}
\begin{center}
\begin{tabular}{p{0.1cm}|p{0.2cm}<{\centering} p{0.4cm}<{\centering} p{0.4cm}<{\centering} | p{1.5cm}<{\centering} p{1.5cm}<{\centering} p{1.5cm}<{\centering}}
\toprule
& NI & DIBR & Pack. &  \textit{30 Scenes}~\cite{DoubleCNN}  &\textit{Reflective}~\cite{StanfordLytro} &\textit{Occlusions}~\cite{StanfordLytro}\\
\midrule
1 & \ding{55} & \checkmark & \checkmark &  40.28 / 0.990 &  37.44 / 0.976 &  34.35 / 0.965 \\
2 & \checkmark & \ding{55} & \checkmark &  36.15 / 0.957 &  37.43 / 0.973 &  32.12 / 0.932 \\
3 & \checkmark & \checkmark & \ding{55} &  39.96 / 0.988 &  37.18 / 0.974 & 33.59 / 0.957 \\
4 & \checkmark & \checkmark & \checkmark & \textbf{40.68} / \textbf{0.990} & \textbf{38.05} / \textbf{0.977} & \textbf{34.54} / \textbf{0.964} \\
\bottomrule
\end{tabular}
\end{center}
\vspace{-4mm}
\end{table}

\subsubsection{Neural interpolation network}
\begin{figure*}
\begin{center}
\includegraphics[width=1\linewidth]{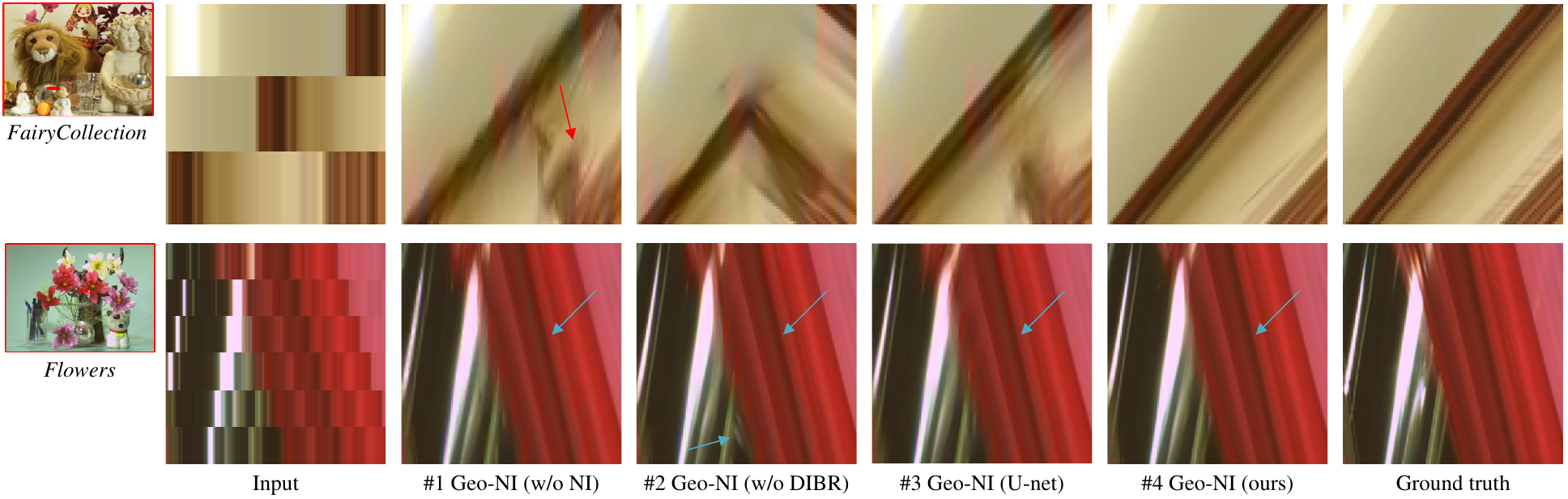}
\end{center}
\vspace{-4mm}
   \caption{Comparison of the proposed Geo-NI framework under different configurations. The LFs are from the MPI Light Field Archive~\cite{kiran2017towards} (top) and the CIVIT Dataset~\cite{ICME2018} (bottom).}
\label{fig:ablation}
\vspace{-3mm}
\end{figure*}

In this experiment, we deactivate the NI network by replacing it with a simplest non-learning reconstruction method, \textit{bilinear} interpolation. This setting is denoted as ``Geo-NI (w/o NI)'' for short, as shown by the setting \#1 in Table~\ref{table:ablations1} and~\ref{table:ablations2}. Benefiting from the perception of scene geometry in the DIBR part, the Geo-NI outperforms the most baseline methods at large downsampling scale (please refer to results of $36\times$ reconstruction in Table~\ref{table:ablations1} and $7\times$ reconstruction in Table~\ref{table:ablations2}), despite it does not use a powerful neural network for LF reconstruction. This ablation study further illustrates the efficacy of the DIBR part in the proposed Geo-NI framework. However, also because of the absence of the reconstruction network, the bilinear interpolation appears an ambiguous result around regions with repetitive texture, as shown in Fig.~\ref{fig:ablation} (top row).

\subsubsection{Depth image-based rendering network}
This setting demonstrates the effectiveness of the DIBR part by detaching all components behind the NI network, which also equals setting the depth hypotheses to zeros. We term this setting as ``Geo-NI (w/o DIBR)'' for short, as shown by the setting \#2 in Table~\ref{table:ablations1} and ~\ref{table:ablations2}. Without using the proposed DIBR part, the Geo-NI (w/o DIBR) is able to perform high-quality reconstruction at a small downsampling scale (please refer to results of $16\times$ reconstruction in Table~\ref{table:ablations1}). But the performances decease significantly when the input LFs become sparser, e.g., the PSNR decreases 3dB for $36\times$ reconstruction on LFs from the MPI Light Field Archive~\cite{kiran2017towards}. As shown in Fig.~\ref{fig:ablation} (top row), the NI network suffers from severe aliasing effects when reconstructing LF with large disparities. On the other hand, it also indicates that the Geo-NI framework is able to enhance the performance of the NI network for LFs with large disparities via the selection of a proper shear amount.

\subsubsection{Packing-unpacking structure}
We validate the effectiveness of the proposed packing-unpacking structure by replacing it with a typical 3D U-net, denoted as ``Geo-NI (U-net)'' for short. Specifically, we discard the S2C-PS (C2S-PS) operations in the packing (unpacking) blocks, and use convolution (deconvolution) layers with stride 2 (0.5) to compress (restore) the spatial information. We also use skip connections between the encoder and decoder instead of those in the residual blocks to transmit the high-frequency components. The dimensions of the network parameters are kept the same to the packing-unpacking network. As shown by the setting \#3 listed in Table~\ref{table:ablations1} and~\ref{table:ablations2}, the Geo-NI (U-net) also achieves a considerable performance with the help of the aliasing measurement mechanism in the DIBR part. However, the Geo-NI with U-net suffers from performance degradation compared with the complete structure (also as shown in Fig.~\ref{fig:ablation}), which demonstrates the effectiveness of the packing-unpacking structure.

\begin{figure}
	\begin{center}
		\includegraphics[width=1\linewidth]{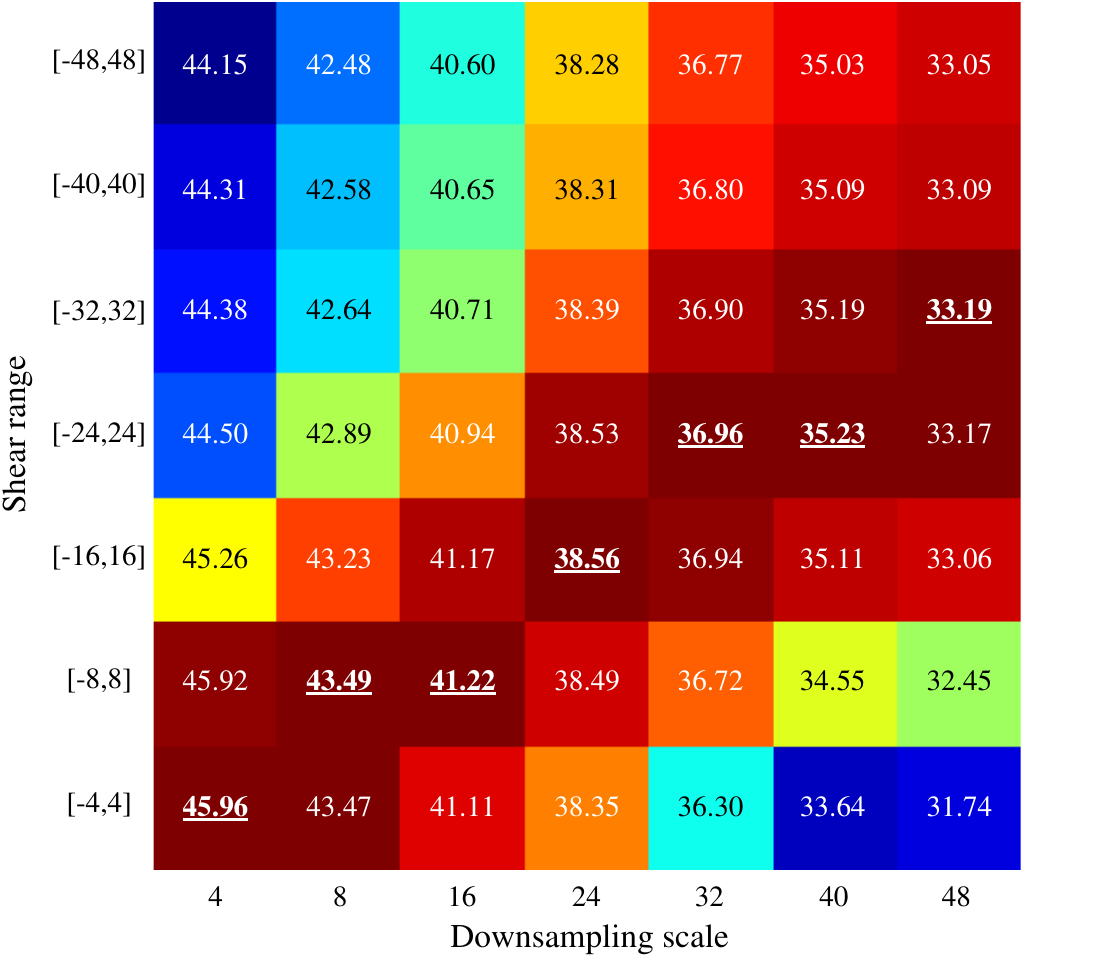}
	\end{center}
	\vspace{-4mm}
	\caption{The performance (PSNR) of the Geo-NI framework for different downsampling scales vs. different settings of shear range. Each grid shows the PSNR value (averaged on LFs from the MPI Light Field Archive~\cite{kiran2017towards}) of a certain downsampling scale under a certain setting of shear range.}
	\vspace{-4mm}
	\label{fig:dispVSshear}
\end{figure}

\subsubsection{Shear range}\label{Sec:ablation_shear}
The proposed Geo-NI framework applies a flexible configurations of shearing without any retraining. In this experiment, we investigate the capability of the Geo-NI framework under different settings of shear range (range of shear amounts). We implement this experiment by evaluating the performance of the Geo-NI under different downsampling scales (in the angular dimension) when choosing different ranges of shear amounts (depth hypotheses), where different downsampling scales will lead to different disparity ranges in the input LF. Note that the network parameters are not retrained in this experiment. The result (evaluated on the MPI Light Field Archive~\cite{kiran2017towards}) is plotted in Fig.~\ref{fig:dispVSshear}, where the horizontal axis denotes the downsampling scale and the vertical axis the shear range. The shear values are evenly sampled every 4 pixels within the range. 

\begin{figure*}
	\begin{center}
		\includegraphics[width=1\linewidth]{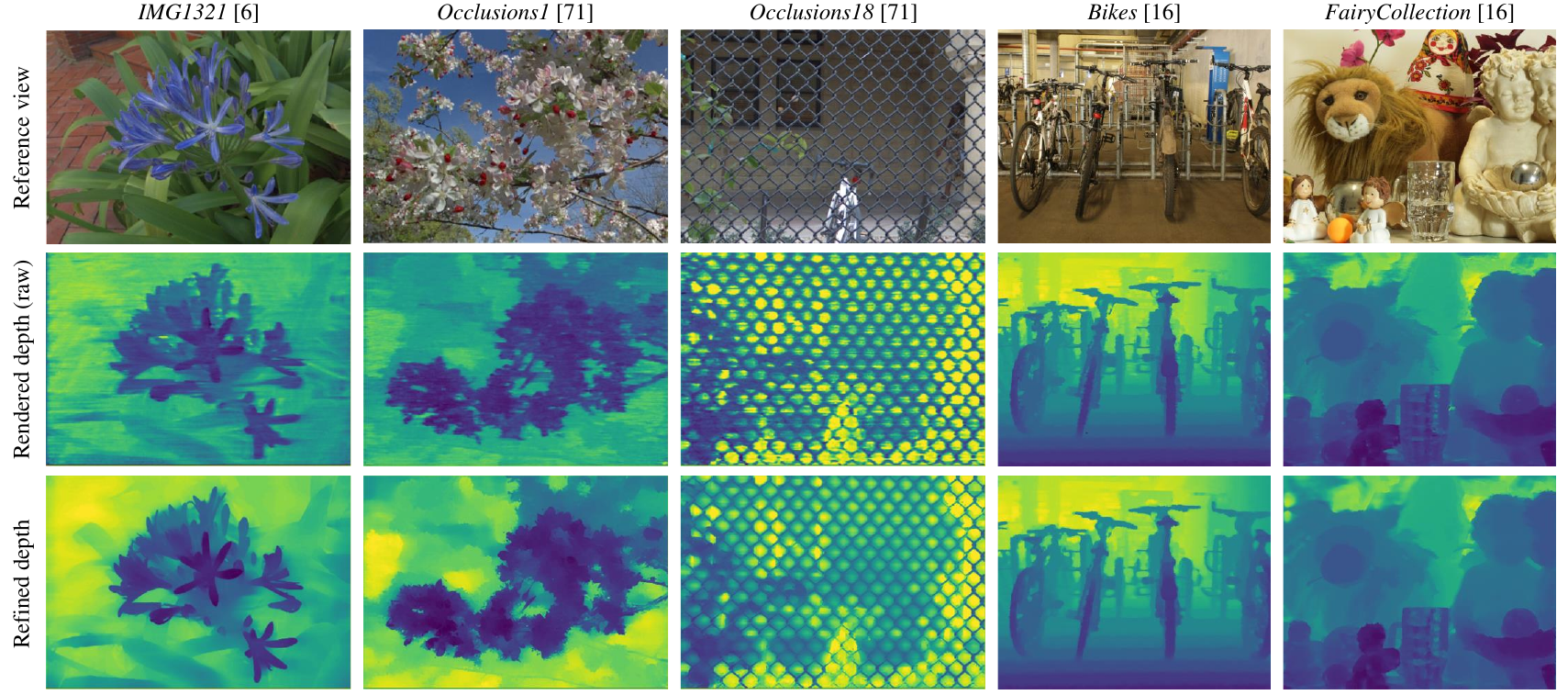}
	\end{center}
	\vspace{-4mm}
	\caption{Illustration of rendered depth maps using reconstruction cost volumes produced by the DIBR network. The middle row shows the depth maps rendered from the raw reconstruction cost volumes. The last row shows the refined depth maps using the classical cost volume filtering method in~\cite{hosni2013costfiltering}.}
	\label{fig:render_depth}
\end{figure*}

The result shows that the proposed Geo-NI is able to achieve a high-quality reconstruction when a reasonable shear range is provided. For instance, for the downsampling scale 48, the disparity range averaged on the test LFs is around $[-30, 35]$, and the Geo-NI provides the best reconstruction results when the shear range is set to $[-32, 32]$. In addition, the result also indicates that the Geo-NI is robust to the settings of the shear range. For example, for the downsampling scale 40 (averaged disparity range $[-25, 30]$), the quantitative result varies from 35.23dB to 35.03dB ($5.7\%$ performance degradation) when we set the shear range from $\pm16$ to $\pm48$. Therefore, the proposed Geo-NI is able to generate satisfactory results even when an optimal shear range is not provided. Fig.~\ref{fig:dispVSshear} also indicates that through adjusting the shear range, the proposed framework is applicable to inputs with different sampling sparsity without retraining.

\section{Further Analysis}\label{Sec:analysis}
In this section, we first introduce the byproduct of the proposed Geo-NI framework, i.e., depth rendering using reconstruction cost volume, then analyse the relation to conventional DIBR and learning-based DIBR.

\subsection{Depth Map Rendering}\label{Sec:depth_render}
The reconstruction cost in the DIBR network implicitly selects a proper shear amount (i.e., depth value) from the hypotheses, which is essentially similar to the purpose of matching cost. Therefore, the DIBR network, to some extent, can be interpreted as a depth estimator. In a standard depth estimation pipeline, the initial depth map can be extracted from the cost volume using the WTA strategy~\cite{scharstein2002taxonomy}, i.e., $\mathop{\arg\min}_{d\in\mathbb{D}}\mathcal{C}_d$. However, this strategy is not able to regress a smooth disparity estimate. In our implementation, we use the soft argmin operation introduced by Kendall~\textit{et al.}~\cite{kendall2017end} to extract the depth information $\mathcal{D}$ as
\begin{equation}\label{eq:depth}
\mathcal{D}(x,y,s)=\sum_{d\in\mathbb{D}}d\cdot\sigma_d(-\mathcal{C}_d(x,y,s)).
\end{equation}
It should be noted that the extracted depth $\mathcal{D}$ is a 3D tensor indicating the variation along the angular dimension.

\begin{figure}
\begin{center}
\includegraphics[width=1.\linewidth]{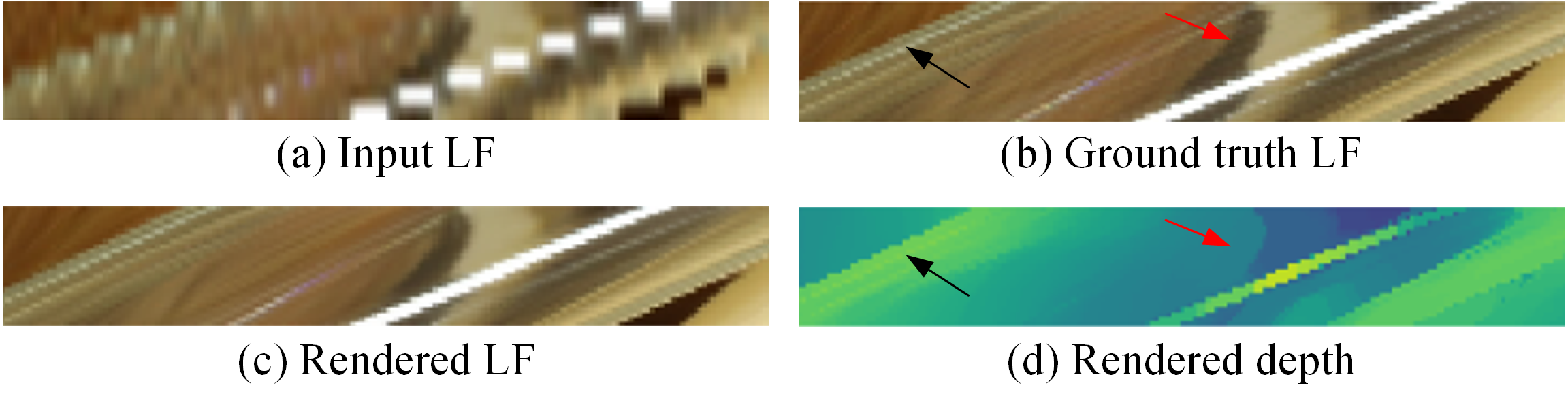}
\end{center}
\vspace{-4mm}
\caption{Measuring the aliasing effect in the reconstructed LF instead of the correspondence in standard cost volume helps to guarantee the photo-consistency for Lambertian regions with large disparities (as shown by the black arrow) as well as the non-Lambertian effect (as shown by the regions around the red arrow).}
\label{fig:aliasing2depth}
\end{figure}

Fig.~\ref{fig:render_depth} shows the rendered depth by using the reconstruction cost volume in the proposed Geo-NI framework. The second and the last rows demonstrate depth maps rendered by using the raw reconstruction cost volumes and the filtered volumes using the classical cost volume filtering method proposed by Hosni~\textit{et al.} in~\cite{hosni2013costfiltering}. The proposed Geo-NI is able to perceive the occlusion boundary despite the training objective of the networks is only LF rendering, as shown by the cases \textit{Occlusion1} and \textit{Occlusion18} in Fig.~\ref{fig:render_depth}.

\subsection{Relation to Depth-Image-based Rendering}\label{Sec:relation2DIBR}
\subsubsection{Relation to conventional DIBR}
Conventional DIBR methods first estimate the depth of the scene, then blend the warped views based on the scene geometry (depth). For the depth estimation, these methods typically construct a cost volume that records the matching cost of each pixel along the dimension of depth hypothesis. The depth of the scene is solved from the cost volume using Winner Takes All (WTA) strategy~\cite{scharstein2002taxonomy}, e.g., argmin operation. Instead of focusing on solving the depth, our idea is to render the desired LF directly from the cost volume. 

\begin{figure*}
\begin{center}
\includegraphics[width=1.\linewidth]{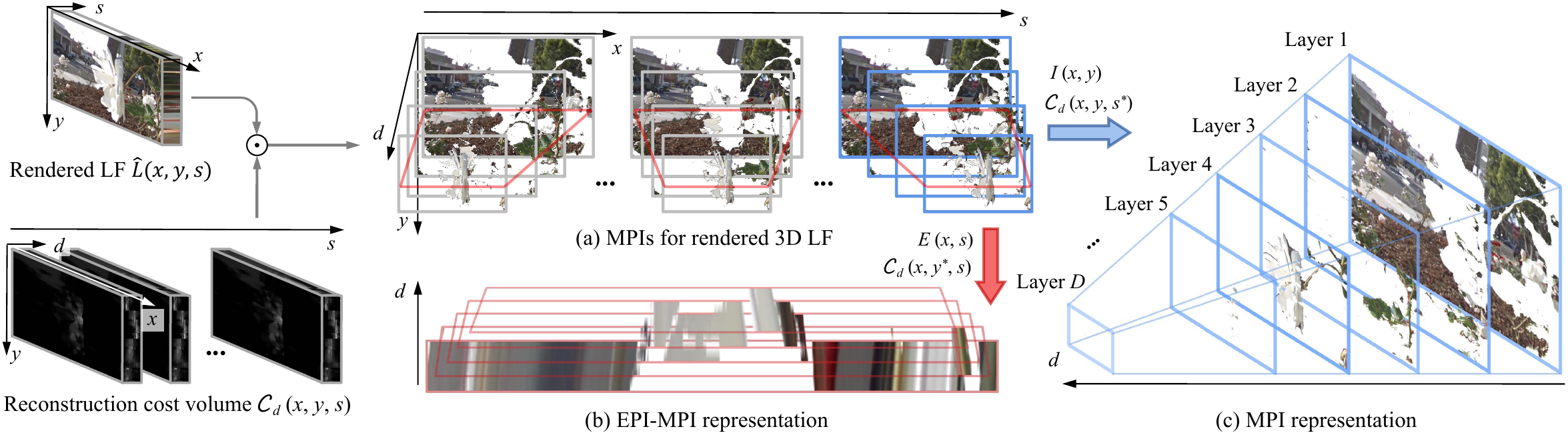}
\end{center}
\vspace{-4mm}
\caption{The reconstruction cost produced by the proposed Geo-NI framework can be interpreted as alpha in the MPI representation. (a) We can promote the rendered 3D LF to MPIs for both input views and reconstructed views. We can also convert the reconstruction cost volume to (b) the EPI-MPI representation or (c) the vanilla version of MPI representation, simply by slicing along different dimensions.}
\label{fig:MPI}
\end{figure*}

But conventional cost volume in DIBR has following problems to reach our idea: i) The cost measures the correspondence between features, it cannot be used to synthesis novel views straightforwardly; ii) The cost is usually aggregated along the angular dimension, leading to dimension mismatching with the LF data. We address these problems by measuring the aliasing effects of the sheared LFs in replacing of the correspondences of a single view. On the one hand, measuring the aliasing effect in the DIBR network helps to guarantee the photo-consistency for Lambertian regions with large disparities, as shown in Fig.~\ref{fig:aliasing2depth} (black arrow). On the other hand, it preserves the non-Lambertian effect reconstructed by the NI network since there is no assumption of depth consistency along the angular dimension, as shown in Fig.~\ref{fig:aliasing2depth} (red arrow).


\subsubsection{Relation to learning-based DIBR}
Standard learning-based DIBR methods~\cite{DoubleCNN,shi2020learning,jin2020learning,jin2020deep,Guo_2021_ICCV} first estimate depth or optical flow via a neural network, and then refine the warped views or feature maps through another network to synthesize the novel view. The refinement network serves to correct the warping errors caused by the depth (flow) estimation network. However, when a region deviates too far from its proper position and the refinement network does not have a large enough receptive field, it will not be corrected. In comparison, the depth perception in the Geo-NI is moved to the rear part of the framework. For regions with small disparities and non-Lambertian effects, the NI network will yield good enough results. For regions with large disparities, the DIBR network is able to select high-quality reconstructed LFs under a certain shear amount through the lens of aliasing.

\subsubsection{Relation to MPI-based Rendering}
MPI-based methods~\cite{zhou2018stereo,mildenhall2019local,John2019DeepView,Srinivasan_2019_CVPR,Tucker_2020_CVPR} decompose input view into multiple fronto-parallel planes at different depths. And the network learns to predict an alpha image at each depth plane. We show that the proposed reconstruction cost can be interpreted as a set of alpha images in the MPI representation simply by slicing the 4D reconstruction cost volume at a certain viewpoint $\mathcal{C}_{d}(x,y,s^*)$, constructing RGB$\alpha$ images  as shown in Fig.~\ref{fig:MPI}(c). But the proposed Geo-NI has the following differences: 1) The proposed Geo-NI framework simultaneously predicts the cost/alpha of both input views and reconstructed views, i.e., the entire 3D LF $\hat{L}(x,y,s)$ (as illustrated in Fig.~\ref{fig:MPI}(a)), promising the view consistency between alpha maps. This feature supports decomposing EPI into an EPI-MPI representation by slicing the 4D reconstruction cost volume at a certain spatial coordinate $\mathcal{C}_{d}(x,y^*,s)$, as demonstrated in Fig.~\ref{fig:MPI}(b). 2) The additional NI network explicitly models the reconstruction of non-Lambertian effects and is able to correct the error caused by false prediction of depth plane. 3) The 3D modeling strategy ensures a flexible number of layers in the MPIs without the need for retraining.

\subsubsection{Relation to reconstruction without implicit depth}
In our previous works~\cite{wu2019learning,wu2021revisiting}, we use a similar strategy that deploys sheared LFs or EPIs in learning-based methods. However, the concept in this paper is fundamentally different. 1) In the proposed framework, we elaborately implant NI within the DIBR pipeline. While in~\cite{wu2021revisiting}, the sheared EPIs is implicitly blended via a fusion network without utilizing the concept of DIBR. The DIBR pipeline in the proposed framework is able to turn the deep neural network into a white box by explaining how the network weights each slice of sheared LF using reconstruction cost. 2) We encode the shear (depth) into the batch dimension of the networks so that we can handle inputs with different sampling by adjusting the shear range without retraining. 3) All the modules in our framework are differentiable, ensuring end-to-end optimization. In contrast, the method in~\cite{wu2019learning} uses non-differentiable post-processing to reconstruct the high-angular resolution LF. 4) The method in~\cite{wu2019learning} requires depth propagation between views before the reconstruction. In the proposed Geo-NI framework, we measure the aliasing effects of the neural interpolated LFs to generate a 3D depth volume (as described in Eqn. \ref{eq:depth}) rather than a single view, which guarantees the view consistency in the rendered LFs.

\subsubsection{Inference speed}
We conduct the statistics of inference speed on an Intel i9-11900K CPU @ 3.50GHz and an NVIDIA RTX 3090 GPU. For LFs of spatial resolution $960\times720$ in the MPI Light Field Archive~\cite{kiran2017towards}, the $16\times$ and $36\times$ reconstructions take 11.16 seconds (0.115 seconds per view) and 18.48 seconds (0.253 seconds per view), respectively. For LFs of spatial resolution $1280\times720$ in the CIVIT Dataset~\cite{ICME2018}, the $16\times$ and $36\times$ reconstructions take 28.46 seconds (0.147 seconds per view) and 59.35 seconds (0.324 seconds per view), respectively. For LFs of spatial resolution $540\times376$ from Lytro Illum, the $7\times$ reconstruction takes 11.36 seconds (0.178 seconds per view). 

\section{Conclusions}
We have proposed a geometry-aware neural interpolation by launching a Neural Interpolation (NI) network within a Depth Image-Based Rendering (DIBR) pipeline. The NI network serves to reconstruct high angular resolution LFs sheared under a set of depth hypotheses. And the DIBR network is developed to construct a reconstruction cost volume through measuring the degrees of aliasing in the neural interpolated LFs, which is then applied for rendering the final LF. We have shown that the reconstruction cost volume in the DIBR network can be used to render scene depth of each view in the LFs or promote the views to MPI representations. This feature brings interpretability to the overall framework. In addition, since we do not compel depth consistency between views, non-Lambertian effects reconstructed by the NI network can be maintained. For the NI and DIBR networks, we have designed a hierarchical packing-unpacking structure that effectively encodes and decodes LF features via spatial-channel pixel shuffling. Evaluations on various LF datasets have demonstrated that the combination of NI and DIBR pipeline is able to render high-quality LFs with large disparities and non-Lambertian effects.

%



%

\ifCLASSOPTIONcompsoc
  \section*{Acknowledgments}
\else
\fi
This work was supported in part by the National Natural Science Foundation of China (NSFC) under Grant 62103092, in part by the Major Program of NSFC under Grants 61991404, and 61991401, in part by the Science and Technology Major Projects of Liaoning Province under Grant 2020JH1/10100008, and in part by the NSFC under Grants U20A20189 and 62161160338.

\ifCLASSOPTIONcaptionsoff
  \newpage
\fi



\bibliographystyle{IEEEtran}
\bibliography{IEEEabrv}
%

%

\begin{IEEEbiography}[{\includegraphics[width=1in]{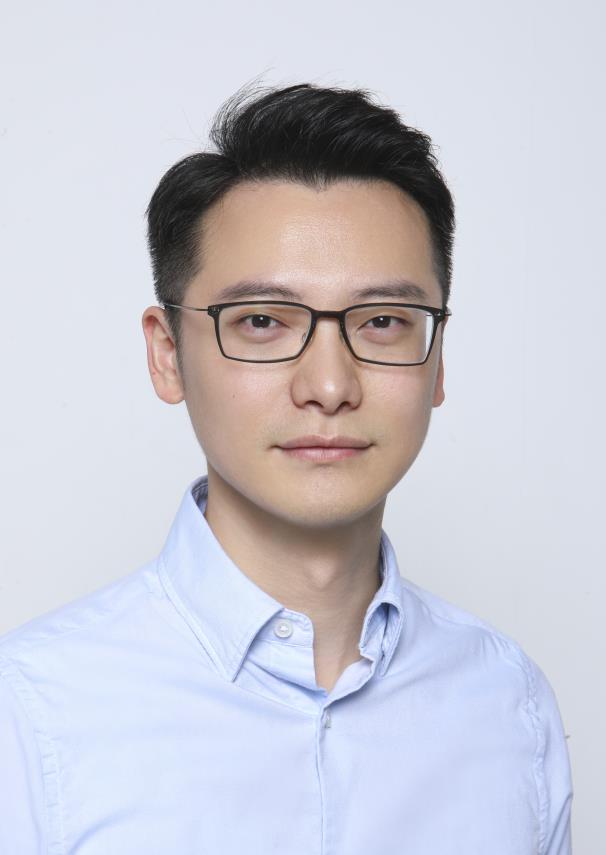}}]{Gaochang Wu}
received the BE and MS degrees in mechanical engineering in Northeastern University, Shenyang, China, in 2013 and 2015, respectively, and Ph.D. degree in control theory and control engineering in Northeastern University, Shenyang, China in 2020. He is currently an associate professor in the State Key Laboratory of Synthetical Automation for Process Industries, Northeastern University. His current research interests include image processing, light field processing and deep learning.
\end{IEEEbiography}

\begin{IEEEbiography}[{\includegraphics[width=1in]{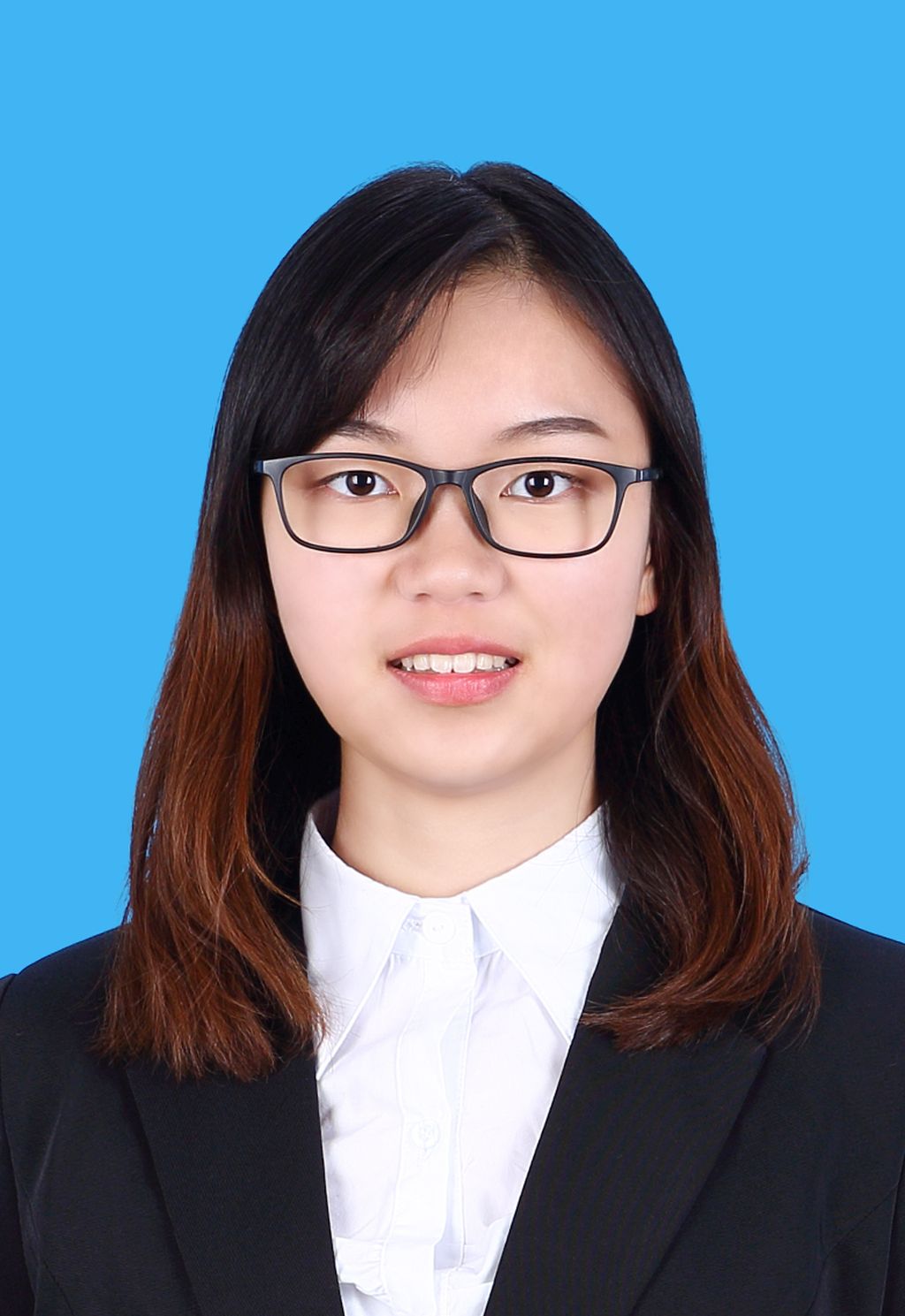}}]{Yuemei Zhou}
received the B.E. degree from Beihang University (BUAA), Beijing, China, in 2018. She is currently pursuing the Ph.D. degree with the Department of Automation, Tsinghua University. Her research interests include image synthesis, image super-resolution and computational photography.
\end{IEEEbiography}

\begin{IEEEbiography}[{\includegraphics[width=1in]{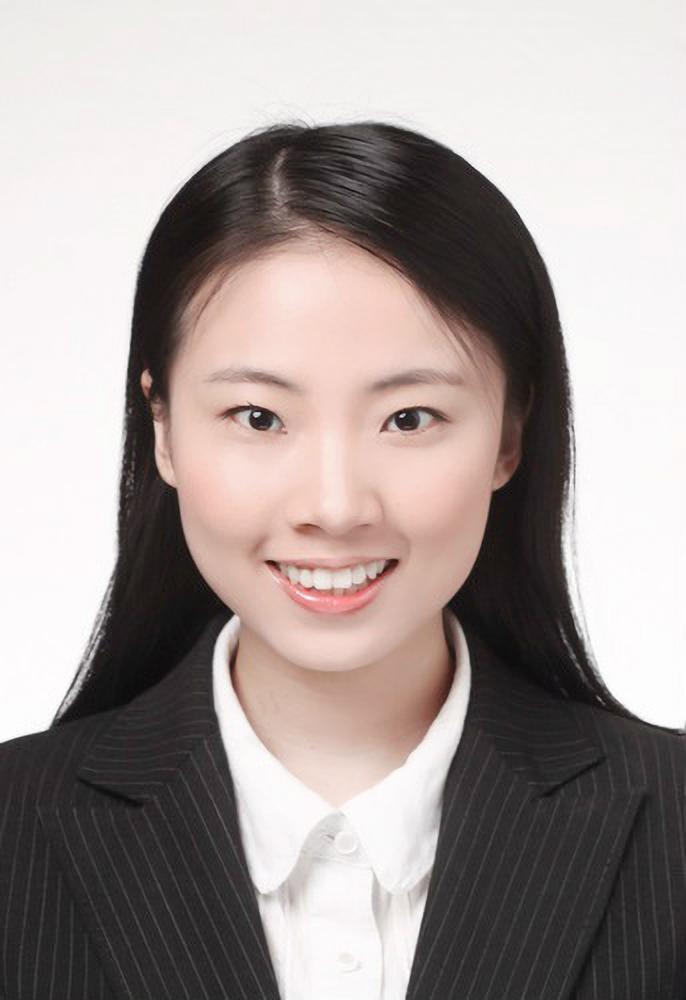}}]{Lu FANG}
is currently an Associate Professor at Tsinghua University. She received her Ph.D in Electronic and Computer Engineering from HKUST in 2011, and B.E. from USTC in 2007, respectively. Dr. Fang's research interests include image / video processing, vision for intelligent robot, and computational photography. Dr. Fang serves as TC member in Multimedia Signal Processing Technical Committee (MMSP-TC) in IEEE Signal Processing Society.
\end{IEEEbiography}

\begin{IEEEbiography}[{\includegraphics[width=1in]{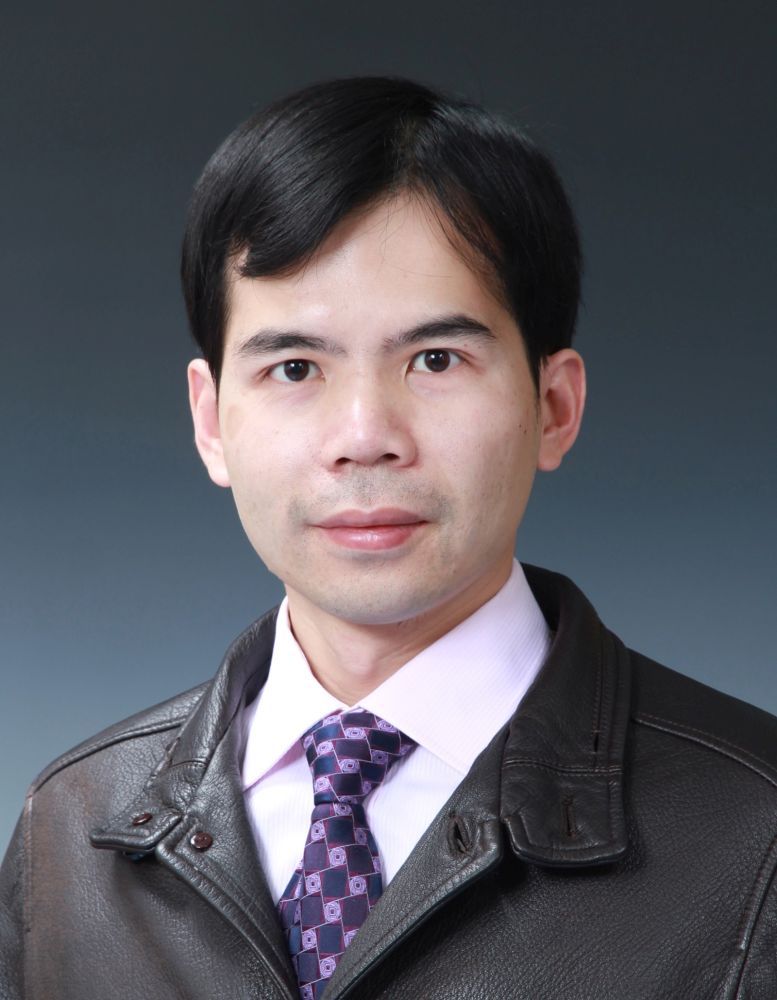}}]{Yebin Liu}
received the BE degree from Beijing University of Posts and Telecommunications, China, in 2002, and the PhD degree from the Automation Department, Tsinghua University, Beijing, China, in 2009. He has been working as a research fellow at the computer graphics group of the Max Planck Institute for Informatik, Germany, in 2010. He is currently an associate professor in Tsinghua University. His research areas include computer vision and computer graphics.
\end{IEEEbiography}

\begin{IEEEbiography}[{\includegraphics[width=1in]{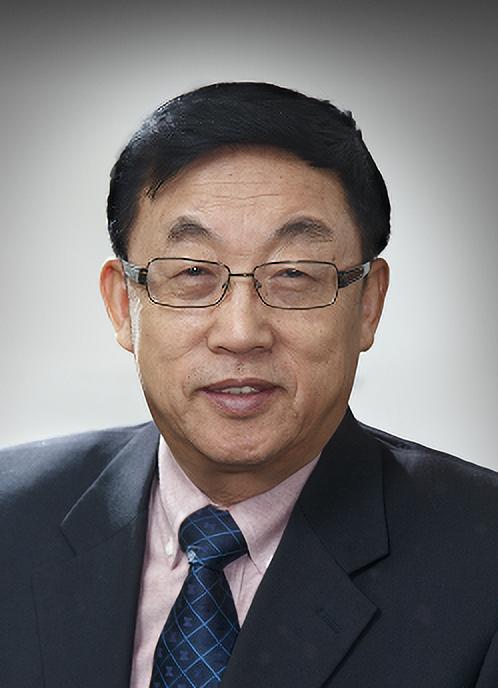}}]{Tianyou Chai}
received the Ph.D. degree in control theory and engineering from Northeastern University, Shenyang, China, in 1985. He has been with the Research Center of Automation, Northeastern University, Shenyang, China, since 1985, where he became a Professor in 1988 and a Chair Professor in 2004. His current research interests include adaptive control, intelligent decoupling control, integrated plant control and systems, and the development of control technologies with applications to various industrial processes. Prof. Chai is a member of the Chinese Academy of Engineering, an academician of International Eurasian Academy of Sciences, IEEE Fellow and IFAC Fellow. He is a distinguished visiting fellow of The Royal Academy of Engineering (UK) and an Invitation Fellow of Japan Society for the Promotion of Science (JSPS).
\end{IEEEbiography}




\end{document}